\pdfoutput=1
\documentclass[11pt]{article}

\usepackage[preprint]{acl}
\definecolor{guidelineblue}{RGB}{0,0,153}
\hypersetup{
  colorlinks=true,
  citecolor=guidelineblue,
  linkcolor=guidelineblue,
  urlcolor=guidelineblue,
  pdftitle={On the Non-Specificity of Statistical Measures Used in Script Decipherment},
  pdfauthor={Nikhil Raghavendra}
}
\usepackage{times}
\usepackage{latexsym}
\usepackage[T1]{fontenc}
\usepackage[utf8]{inputenc}
\usepackage{microtype}
\usepackage{inconsolata}
\usepackage{graphicx}
\usepackage{amsmath,amssymb}
\usepackage{booktabs}
\usepackage{array}
\newcolumntype{L}[1]{>{\raggedright\arraybackslash}p{#1}}
\usepackage{float}
\usepackage{etoolbox}
\usepackage{placeins}

\AfterEndEnvironment{figure}{\FloatBarrier}
\AfterEndEnvironment{figure*}{\FloatBarrier}

\graphicspath{{figures/}}
\newcommand{\sigil}{\textsc{Sigil}}

\title{On the Non-Specificity of Statistical Measures Used in Script Decipherment}
\author{Nikhil Raghavendra \\
  Independent Researcher \\
  \texttt{me@nikhilr.io}}

\begin{document}
\maketitle

\begin{abstract}
Statistical regularities are routinely offered as evidence that undeciphered
sign systems encode language; the Indus script debate is the canonical
example. Any such inference rests on specificity: the reported
outcome must be unusual among plausible structured non-languages. We test that
premise constructively with \sigil{}, a purpose-built generative emblem
system whose 3,000-text core corpus carries explicit compositional meanings
although no sign has a phonological value. A literature registry compiled in
advance of evaluation records 54 methods and admits a method to exact scoring when both the published
Indus outcome and a source-defined decision rule can be reproduced. \sigil{}
receives the same category as the Indus corpus on every criterion scored this
way, across repetition, directional-asymmetry, and lexical-distribution
tests. Declared reconstructions of entropy,
frequency, positional, predictive, classifier, and network measures reproduce
the familiar Indus-like signatures as well. A sequential decipherment stress
test then reaches high dictionary coverage for English, Sanskrit, and Tamil
on the same corpus, while grouped held-out declines and unstable keys reveal
how little that coverage identifies. The construction does not decide what the Indus
signs encode: it shows that the evaluated measures detect organization
without being specific to language, and therefore cannot, on their own,
establish encoded speech.
\end{abstract}

\section{Introduction}\label{sec:intro}

When a script resists decipherment, statistics step into the space that
bilingual texts would normally occupy. Conditional entropy has been used to
argue that the Indus signs behave like writing \citep{rao2009science}, an
entropy-based classifier to argue that Pictish symbols record a language
\citep{lee2010pictish}, and a recent cryptanalytic proposal reads the Indus
corpus as Sanskrit on the strength of dictionary-supported pattern matching
\citep{yajnadevam2024}. The stakes are real: the government of Tamil Nadu
announced a one-million-dollar prize for a decisive decipherment in January
2025 \citep{indiatoday2025}, and claimed solutions attract wide attention
\citep{gordon2026}. This paper asks what such statistical evidence can and
cannot establish.

The Indus civilization left several thousand short inscriptions on seals,
sealings, tablets, and small objects. Mahadevan's concordance catalogued
2,906 inscribed artifacts \citep{mahadevan1977}, later documentation has
continued to expand the corpus \citep{wells2015icit}, and a typical text
carries fewer than five signs. No bilingual text is known, proposed
readings disagree about language and sign function \citep{parpola1994},
and a prominent line of scholarship disputes whether the corpus is writing
at all \citep{farmer2004}, itself contested \citep{vidale2007}.
A substantial literature has therefore measured the corpus instead, reading
conditional and block entropy, rank-frequency laws, positional restrictions,
predictability, network structure, and classifier output as signs of language
\citep{rao2009science,rao2010ieee,yadav2010plos,sinha2011network}. These are
genuine properties of the data; the difficulty lies in the comparison class,
because administrative and emblematic practices can also generate skewed
frequencies, recurring slots, and local dependencies, and rigid or
independently random controls do not represent that space
\citep{sproat2010cl,sproat2014language}.

We therefore ask a narrower question than decipherment: are the published
statistical outcomes specific enough to diagnose language? Our answer is a
constructive counterexample. \sigil{} is a deterministic generative system
that produces seal-like inscriptions for a fictional administrative world.
Every text has meaning, because every text records a dedication, an office,
an entitlement, or an administrative event whose semantic derivation is
stored beside it; no text has linguistic encoding, because no sign carries
a sound, word, or morpheme value and no inscription transcribes an
utterance. The generator, its parameters, and the aligned semantics are
public, so this negative fact is inspectable rather than asserted.

Against a method-level registry closed before evaluation began, the
canonical \sigil{} corpus reproduces every source-validated categorical outcome
that the registry admits to exact scoring, and declared reconstructions
place it in the familiar Indus-like regions of the standard entropy,
frequency, positional, predictive, classifier, and network analyses. A
further experiment turns to decipherment itself: optimized sign-to-letter
keys make most of the corpus segmentable into English, Sanskrit, or Tamil
dictionary forms, three mutually incompatible readings of signs that encode
nothing, while grouped held-out evaluation and key instability reveal how
weakly each individual value is identified.

The claim we defend is one of constructive existence. A single explicit
non-language that satisfies the tested criteria refutes the proposition
that those criteria logically require encoded language. It does not
estimate a likelihood ratio, show that the measures carry no evidential
value under every calibrated model, or determine what the Indus signs
encode. What it does change is the burden of proof, for the Indus debate
and for statistical decipherment arguments generally.

\section{Specificity as the Inferential Target}\label{sec:specificity}

Let $T_m(x)$ denote the outcome of method $m$ on corpus $x$. A categorical
criterion could establish language over a stated comparison class
$\mathcal{C}$ only if every $x\in\mathcal{C}$ receiving the
language-associated outcome were linguistic, so one explicit member of
$\mathcal{C}$ known to be non-linguistic disproves that implication for the
chosen class, although it says nothing about prevalence. In probabilistic
terms, the construction establishes $P(T_m\mid N)>0$ for at least one fully
specified non-language model $N$; it does not determine
$P(T_m\mid L)/P(T_m\mid N)$ over broader families, which is what a
calibrated evidential argument would need.

The distinction matters because description and diagnosis are different
tasks. Conditional entropy, Zipf-Mandelbrot fits
\citep{zipf1949,mandelbrot1953}, and transition graphs each reveal that a
corpus is organized without identifying the substance being organized, and
what counts as writing was never a purely statistical question to begin
with \citep{gelb1952}. The null hypothesis then fixes what a rejection
means: shuffling signs within texts tests whether the observed order
differs from that permutation, and it cannot establish that the order comes
from syntax rather than ritual sequence, administrative slots, or
production habit.

Earlier critiques pressed this point with i.i.d. processes and historical
non-writing corpora \citep{liberman2009,sproat2010cl,sproat2014language}.
\sigil{} adds what those arguments lacked: a fully specified generative
witness whose origin is known, whose meanings are aligned text by text, and
whose parameters are public. The witness is intentionally favorable to the
statistical criteria, which is what a counterexample to sufficiency should
be, although it thereby rules out any claim of held-out prediction or
population-level calibration on our part.

\section{How Statistical Evidence Entered the Debate}\label{sec:literature}

The modern round of the debate opened with a denial. \citet{farmer2004}
argued that the Indus corpus, with its very short texts and its scarcity
of long repeated sequences, records no language at all, and that a
century of decipherment attempts had pursued a literate civilization
that never existed. The statistical case for language grew up largely in
answer to that challenge, accumulating method by method through analyses
of uncertainty, frequency, order, and graph structure whose results are
now often cited together. Cited in bulk, a varied literature can look
like one convergent experiment, even though the studies differ in
corpus, orientation, filtering, control data, and inferential target.

The most visible answer came five years later in a general-science
venue. \citet{rao2009science} estimated bigram conditional entropy while
growing the retained vocabulary and found the Indus curve between
selected natural-language references and the more rigid or less
constrained controls, and follow-up work extended the argument to block
entropy \citep{rao2010ieee,rao2010cl}. A companion Markov analysis
reported non-uniform initial signs, restricted successors, and far lower
likelihoods for inscriptions found in West Asia than for held-out
core-region material, which was read as a flexible sign system adapted
abroad \citep{rao2009pnas}. These studies establish that the corpus is
neither a fixed template nor an independent equiprobable process.

Objections arrived quickly. \citet{liberman2009} argued that the
intermediate entropy band is easy for non-linguistic processes to enter,
\citet{sproat2010cl} showed that an independent Zipfian source occupies
the same band and faulted the reviewing practices that had admitted the
claim, and replies and rejoinders followed on both sides
\citep{rao2010cl,lee2010cl,sproat2010reply}.

The descriptive toolkit widened anyway. \citet{yadav2010plos} reported a
Zipf-Mandelbrot fit, a sharp asymmetry between the common initial and
common final signs, a large unigram-to-bigram perplexity drop, and
successful restoration of deleted signs, while earlier segmentation work
had decomposed longer inscriptions using repeated complete texts and
frequent combinations \citep{yadav2008segmentation}. All of this is
consistent with combinatorial sign use, yet none of it says whether the
recurring units are words, titles, commodities, deities, or other emblem
classes.

Classification and network analyses then made the linguistic reading
explicit. \citet{lee2010pictish} embedded small symbol corpora in a
decision plane built from unigram and bigram features and labeled its
regions with terms such as ``writing: syllables,'' a classification that
drew immediate criticism of its own \citep{liberman2010}.
\citet{sinha2011network} compared transition-graph connectance and
reciprocity with within-text shuffles and interpreted the differences as
syntax-like. Directional asymmetry was later inferred from the
concentration and entropy of terminal signs \citep{ashraf2018handedness},
and LNRE models brought formal goodness-of-fit tests to the frequency
spectrum \citep{oakes2019}.

The dissent returned with an instrument of its own.
\citet{sproat2014language} proposed an adjacent-to-total repetition
ratio on which the Indus corpus falls among his non-linguistic systems,
itself contested further \citep{rao2015language,sproat2015reply}. Most
recently, a self-published proposal combines dictionary-supported
regular expressions with a borrowed unicity-distance argument to claim a
Sanskrit reading \citep{yajnadevam2024}, raising a different question:
whether a large lexical search space can make a short corpus readable
regardless of what it records.

Our registry preserves these distinctions instead of flattening them. A
statistic can be computed correctly while its interpretation remains
disputed, and a source can define a categorical outcome that is explicitly
non-linguistic. We therefore score reproducible outcomes exactly as their
authors defined them and treat continuous resemblance as description, which
keeps the literature from becoming a vote whose members were never designed
to be exchangeable.

\section{A Non-Linguistic System With Meaning}\label{sec:sigil}

\begin{figure}[!htb]
\centering
\includegraphics[width=\columnwidth]{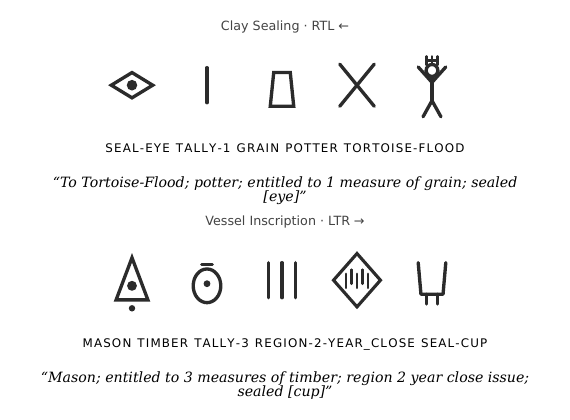}
\caption{Two artifacts from the canonical \sigil{} corpus, one for each
reading direction: a clay sealing read right to left and a vessel
inscription read left to right. Each row shows the physical surface, the
reading direction, the sign glosses, and the generated compositional
meaning. The italic readings are mechanical renderings of the stored
semantics, not decipherments.}
\label{fig:seals}
\end{figure}

\paragraph{The fictional practice.}
Figure~\ref{fig:seals} shows the system at work. A seal can name a patron
deity, attach licensed epithets, identify a guild and a rank within it,
list commodity entitlements with tally marks, and close with a validating
emblem. Four core regions share a pantheon, each with a designated patron
and its own slot conventions, and a trading outpost mixes the shared
deities with two local cults under a looser ordering convention. Twenty short-lived
administrative issue marks are stamped onto bounded sets of clay sealings
and vessel inscriptions, so the corpus contains meaningful rare signs whose
rarity is administrative rather than phonological.

The design draws on the broad archaeological record of what seals and
emblems have done in literate and non-literate settings alike, marking
ownership, institutional authority, cultic affiliation, and administrative
control from Mesopotamian sealing practice to Anatolian pot marks and the
Gulf trade
\citep{collon1987,zettler1987,englund2011,glatz2012,laursen2010,ameri2013},
and proposals that the Indus signs conveyed meaning through largely
non-phonological conventions have been advanced in the recent literature
\citep{mukhopadhyay2019}. \sigil{} remains fictional throughout: its labels
and glyphs claim no affinity with any attested language, polity, or signary,
and no statistical method in this paper inspects glyph geometry.

\paragraph{Composition.}
Each inscription is assembled from optional opener, dedication, office,
ledger, and terminal blocks. Finite-pool Pitman-Yor urns govern the dynamic
categories, with a concentrated head and a long tail of deities, epithets,
guilds, and commodities \citep{pitman1997}. Deities license weighted
multisets of epithets, so particular deity and epithet combinations recur
far above chance; commodity entries may carry tally marks or an emphatic
doubling; after the first 1,200 texts, seventeen frequent adjacent pairs
become registered ligatures that later texts may write as fused signs; and
a small probability permits adjacent blocks to invert. These mechanisms
create local constraint and controlled variation while leaving the semantic
role of every sign explicit.

Material support and orientation are drawn from independent random streams.
Region-specific reinforced urns assign 1,100 seal matrices, 1,475 clay
sealings, and 425 vessel inscriptions across the core corpus; matrices are
cut in mirror order for a right-to-left impression, and vessels draw their
direction from a reinforced urn favoring right-to-left by nine to one, so
the canonical reading directions are 2,972 right-to-left and 28
left-to-right. Sequence analyses consume the semantic reading order, while
the handedness procedure consumes a separately stored normalized spatial
view, exactly as its source method requires.

\begin{table}[t]
\centering
\small
\begin{tabular}{ll}
\toprule
Canonical \sigil{} quantity & Value \\
\midrule
Core texts / outpost texts & 3,000 / 350 \\
Core tokens / attested types & 13,717 / 419 \\
Distinct core strings & 2,753 \\
Registered signs / ligatures & 450 / 17 \\
Mean / maximum core length & 4.572 / 12 \\
Signs covering 80\% of tokens & 51 \\
Top-sign token share & 6.99\% \\
Right-to-left / left-to-right & 2,972 / 28 \\
\bottomrule
\end{tabular}
\caption{The released fixed-seed corpus. All counts come from the portable
corpus artifact and the current result contract.}
\label{tab:corpus-main}
\end{table}

\paragraph{The released corpus.}
Table~\ref{tab:corpus-main} summarizes the canonical realization. One
pinned seed controls composition, and three further seeds independently
control material support, vessel direction, and administrative issues. The
generator runs once, its sign registry becomes immutable after generation,
and every text carries aligned semantic, physical-surface, and
normalized-spatial records. The 419 attested core types are drawn from 450
registered identities, and fifteen further types occur only at the outpost.

The corpus is purpose-built, and we present it as such: its size and
inventory sit in the range used by the Indus studies, and its mechanisms
were chosen with the published statistics in view, because the question
under test is precisely whether a known non-language can satisfy their
stated outcomes. To check that the headline behavior is not an accident of
one
parameter point, we also evaluate a common-seed grid of 27 configurations
varying urn concentration, optional-slot probability, and epithet-license
length, a sensitivity display and not a variance estimate.

\section{The Evaluation Protocol}\label{sec:methods}

\subsection{A method-level registry}

The literature census closed before any scoring ran.
Its 54 entries represent methods rather than papers or individual plotted
values, so one publication may contribute several entries, and each entry
records the source corpus, orientation, filtering, formula, control
construction, randomization, tie handling, required inputs, and the status
of our reproduction. The census contains 6 exact-scored methods, 18
reconstructions, 14 descriptive entries, 10 blocked methods, and 6
exclusions.

An entry reaches the exact denominator only when two conditions hold
together: we must independently reproduce the reported Indus outcome from
public source material, and the publication must state a categorical
decision or null-test rule that can be applied without inventing a
similarity tolerance. Continuous resemblance is never scored. A
reconstruction records a declared implementation of a method whose source
leaves consequential details open, a blocked entry lacks a necessary
resource, rule, or defensible target, and an exclusion falls outside the
registry's scope. These statuses keep every unavailable method visible, so
that missing information can never quietly become a favorable result.

For an eligible method, a pass means only that
$I_m=\mathbb{1}[T_m(\text{\sigil{}})=T_m(\text{Indus})]$, and the aggregate
is the fraction of eligible methods with $I_m=1$. The aggregate is not a binomial test and we do not treat
it as one: two entries come from a single handedness procedure, three are
related LNRE goodness-of-fit models, and one is a repetition ratio, so
their mutual dependence is substantive and acknowledged.

\subsection{Descriptive reconstructions}

The broader battery covers the measurements that recur throughout the debate.
It includes modified Kneser-Ney conditional entropy \citep{kneser1995,chen1998},
NSB block entropy \citep{nsb2002}, Zipf-Mandelbrot fitting, positional
concentration, Witten-Bell perplexity, masked-sign restoration, $n$-gram
counts, Dunning bigram collocations \citep{dunning1993}, staged segmentation,
the Lee decision rule, transition networks, outpost likelihoods, terminal
directionality, and LNRE models. Appendix~\ref{app:methods} states the
operational choices and the departures from each source. All of these
reconstructions are evidence about compatibility; none of them enlarges the
exact score.

The controls include a rigid repeated template, an equiprobable
maximum-entropy process, an independent Zipf-Mandelbrot process, within-text
shuffles of \sigil{}, and pinned Brown-corpus word and character sequences
\citep{francis1979}. Vocabulary filtering preserves the gaps it creates,
and null figures plot the stored permutation draws rather than samples
synthesized from their moments.

\begin{table*}[t]
\centering
\small
\renewcommand{\arraystretch}{1.35}
\begin{tabular}{L{.19\textwidth}L{.22\textwidth}L{.22\textwidth}L{.25\textwidth}}
\toprule
Method & Source-defined rule & Reproduced Indus outcome & \sigil{} outcome \\
\midrule
Adjacent / total repetition
& $r/R<0.10$ linguistic; $>0.10$ non-linguistic
& $23/40=0.575$, non-linguistic
& $283/438=0.646$, non-linguistic \\
Terminal Gini asymmetry
& positive estimate, BCa excludes zero, shuffle $p<0.05$
& $\Delta G=0.1788$, RTL, $p=0.001$
& $\Delta G=0.0920$, RTL, $p=0.001$ \\
Terminal entropy asymmetry
& negative estimate, BCa excludes zero, shuffle $p<0.05$
& $\Delta S=-0.4144$, RTL, $p=0.001$
& $\Delta S=-0.2920$, RTL, $p=0.001$ \\
GIGP LNRE fit
& chi-square goodness-of-fit $p<0.05$
& reject, $p=6.51\times10^{-4}$
& reject, $p=5.73\times10^{-4}$ \\
Finite ZM LNRE fit
& chi-square goodness-of-fit $p<0.05$
& reject, $p=1.28\times10^{-9}$
& reject, $p=1.40\times10^{-6}$ \\
ZM LNRE fit
& chi-square goodness-of-fit $p<0.05$
& reject, $p=1.23\times10^{-23}$
& reject, $p=4.90\times10^{-7}$ \\
\bottomrule
\end{tabular}
\caption{The complete exact denominator. A match means the same categorical
or null-test outcome under the source-defined rule. It does not mean
numerical identity, independent evidence, or support for a linguistic
reading. ZM abbreviates Zipf-Mandelbrot and RTL right-to-left.}
\label{tab:exact}
\end{table*}

\subsection{Sequential lexical stress test}\label{sec:decoder-method}

The recent Sanskrit proposal treats repeated-sign regular expressions,
intersections of candidate values, and dictionary-supported segmentation as
evidence for a reading \citep{yajnadevam2024}. Its public implementation
contains manually written expressions for 32 sign headings, but it
specifies no complete signary algorithm, generic search schedule, beam
policy, segmentation procedure, held-out design, or tie rules. We therefore
classify our version as a source-constrained reconstruction rather than a
replication, and we apply it to \sigil{} as a stress test of what
dictionary coverage alone can show.

A single deterministic engine built from equality patterns, candidate
intersection, a small beam, and dynamic-programming segmentation is applied
to English, Sanskrit, and Tamil. Each sign maps to one alphabet unit,
homophony allows several signs to share a unit, repeated signs impose
back-reference constraints, and whole-word constraints from short or
repeated inscriptions restrict candidate values, after which a declared
lexicographic objective drives beam search and coordinate refinement.
Dynamic programming then asks whether each rendered string can be segmented
completely into pinned dictionary forms.

The literal protocol changes nothing beyond homophony and complete
segmentation, and it is the only protocol English and Tamil receive. Sanskrit
additionally has a script protocol with declared sound-class folds, optional
inherent \emph{a}, consonant-only realizations, mechanical inflection and
sandhi, and decomposition of the 17 genuine ligatures, together with an
adaptive level that permits a bounded, logged overlay and downstream key
revision. The levels are cumulative and always reported separately.

Exact duplicate inscriptions stay inside a single deterministic 70/15/15
train, validation, or test group. The source-style result optimizes and
scores the full corpus, while the grouped result trains a fresh key, and
any adaptive additions, on training groups alone. Twenty random keys,
fixed-key shuffled and independent controls, four deterministic starts, six
bootstrap keys, complete key tables, and segmentation multiplicities are
all stored in the released contract, and \sigil{} meanings never enter
selection, key search, segmentation, or scoring.

\section{Results}\label{sec:results}

\subsection{The exactly scored outcomes coincide}

Table~\ref{tab:exact} presents the exactly scored comparisons, and the
parallels run closer than the category labels alone suggest. \sigil{}'s
adjacent-to-total repetition ratio of 0.646 sits beside the reproduced
Indus value of 0.575, both far above the 0.10 boundary the source
associates with linguistic systems. The two directional criteria recover
\sigil{}'s physical right-to-left convention just as they recover the
Indus convention, with matching signs on both asymmetries and the same
permutation significance of $p=0.001$, and the three LNRE models are
rejected for both corpora, with the two GIGP rejections landing at the
same order of magnitude.

The reading of this agreement is nonetheless precise and limited. The
repetition criterion places both corpora in its non-linguistic category,
the directional criteria measure handedness without measuring language,
and an LNRE rejection establishes poor fit to those lexical models, not
membership in a linguistic class. Every source-validated categorical
outcome in the registry is compatible with an explicit non-language.
Reading the rows as independent language tests, or the agreement as a
calibrated probability, would overstate what the registry supports.

\subsection{The descriptive signatures also recur}

\begin{figure}[!htb]
\centering
\includegraphics[width=\columnwidth]{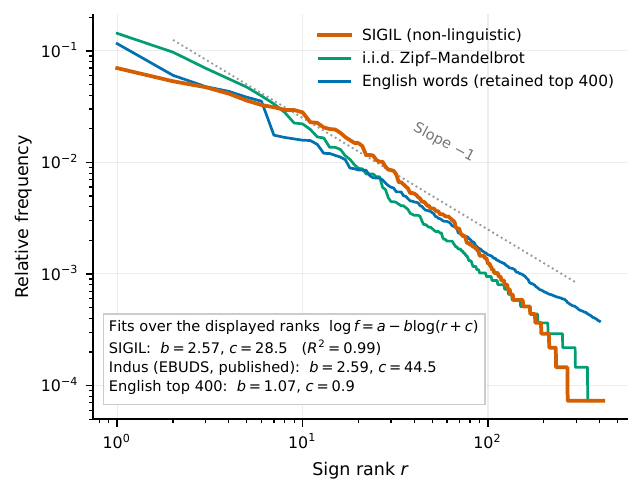}
\caption{Rank-frequency curves and fitted Zipf-Mandelbrot models. \sigil{}
has $b=2.574$, $c=28.46$, and $R^2=0.986$ over the displayed ranks; the
published EBUDS parameters are $b=2.59$ and $c=44.47$. The English fit
uses the retained top-400 word corpus.}
\label{fig:zipf}
\end{figure}

\paragraph{Frequency concentration.}
Figure~\ref{fig:zipf} gives the first descriptive comparison. The \sigil{}
curve pairs a concentrated head with a long tail of rare signs, and its
fitted Zipf-Mandelbrot exponent of 2.574 sits close to the published EBUDS
value of 2.59, with $R^2=0.986$ over the fitted ranks. The mid-rank log-log
slope is $-1.270$, against about $-1.03$ for the displayed English-word
reference. Category-specific reinforced urns produce this shape as readily
as word use does, so the fitted distance receives no categorical score.

\begin{figure}[!htb]
\centering
\includegraphics[width=\columnwidth]{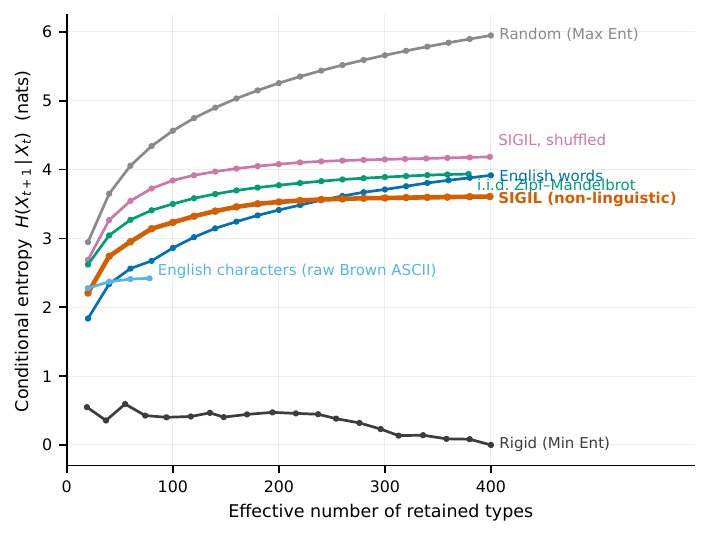}
\caption{Smoothed bigram conditional entropy in nats as the retained
vocabulary grows. \sigil{} lies between the Brown word and character
references and below its own within-text shuffle. The i.i.d.
Zipf-Mandelbrot control has no sequential dependence yet also occupies an
intermediate region.}
\label{fig:ce}
\end{figure}

\paragraph{Sequential uncertainty.}
Figure~\ref{fig:ce} shows the conditional-entropy sweep that anchors the
entropic argument. At the ends of their sweeps, \sigil{} reaches 3.609
nats, English characters 2.421, English words 3.917, shuffled \sigil{}
4.185, and the equiprobable control 5.948. The constructed system occupies the intermediate band for a simple
reason: its next sign is constrained by slots and licenses but never fixed. The
independent Zipf-Mandelbrot control occupies a neighboring band with no
sequential structure at all, which is the objection earlier critics raised
against the entropic argument, reproduced here inside one controlled
pipeline.

\begin{figure}[!htb]
\centering
\includegraphics[width=\columnwidth]{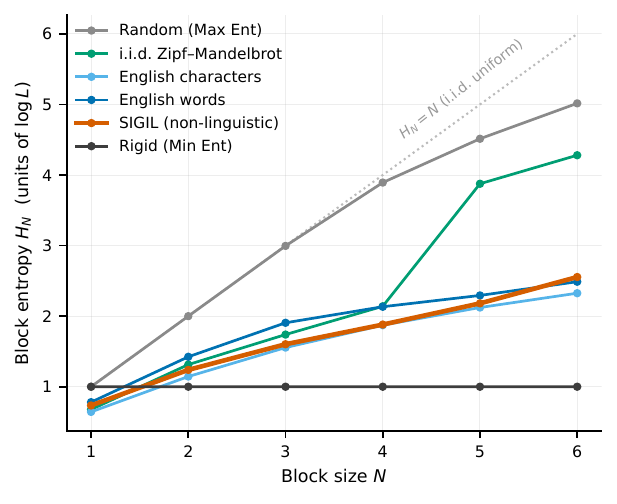}
\caption{NSB block entropy in units of $\log L$, with $L$ the alphabet
size. \sigil{} grows sublinearly, staying far below the equiprobable
control at every order. The late rise of the i.i.d. control occurs where
almost every six-sign block in that corpus is unique.}
\label{fig:block}
\end{figure}

\paragraph{Longer blocks.}
Figure~\ref{fig:block} extends the comparison to block entropies of orders
one through six. The six \sigil{} values are 0.734, 1.239, 1.602, 1.881,
2.181, and 2.554, while English words reach 2.488 at order six and English
characters 2.325. Because the corpus consists of short texts with repeated
templates, these estimates reflect text length and template reuse alongside
any sequential dependence, so location within the linguistic reference region
cannot function as a diagnostic.

\paragraph{Prediction and restoration.}
Sequence prediction behaves the same way as the entropies. Five-fold
Witten-Bell perplexity falls from 64.22 at order one to 25.80 at order two
and then ranges between 28.59 and 30.39 through order five, while the
published Indus sequence runs 68.82, 26.69, 26.09, 25.26, and 25.26.
The mild rise beyond order two reflects the 4.572-sign mean text length,
which leaves few long contexts for higher orders to exploit; the
diagnostic drop from order one to order two appears in both corpora.
Restoration behaves similarly: when one sign is deleted, the correct value
lies inside the smallest candidate set holding 90\% of posterior mass in
84.88\% of 275,300 trials, against a published Indus figure of 74\% for
the same criterion, and top-one accuracy is 24.84\%. Bigram constraint with little higher-order gain is
what slot composition of short texts produces.

\begin{figure}[!htb]
\centering
\includegraphics[width=.88\columnwidth]{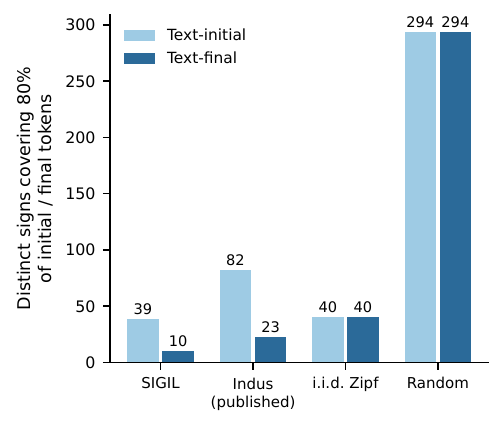}
\caption{Numbers of distinct signs needed to cover 80\% of text-initial and
text-final tokens. Slot structure gives \sigil{} a marked asymmetry between
beginners and enders, while the i.i.d. and equiprobable controls remain
approximately symmetric.}
\label{fig:position}
\end{figure}

\paragraph{Position.}
Figure~\ref{fig:position} compares positional concentration. Thirty-nine
\sigil{} signs cover 80\% of initial tokens while ten cover 80\% of final
tokens, a 39-to-10 contrast that parallels the published 82-to-23 Indus
asymmetry. Openers, dedication and office blocks, issue marks, and a
five-sign terminal class generate this structure in \sigil{}. The statistic
detects a genuine slot system in both corpora; what it cannot say is
what the slots hold.

The graph and classifier summaries complete the profile. The transition
graph has 419 vertices, 3,209 directed edges, and 51 self-loops; its
reciprocity of 0.145 falls far below the within-text-shuffle mean of 0.467,
just as the published Indus values of 0.148 and 0.338 do, and its connectance
of 0.0183 falls below the null value of 0.0264, with published Indus values
of 0.0077 against 0.011. The Lee rule assigns \sigil{} to ``writing:
syllables'' at $(C_r,U_r)=(8.93,1.33)$, and all 27 sensitivity
configurations retain that label \citep{lee2010pictish}.

The outpost experiment parallels the West Asia design where the truth is
known
\citep{rao2009pnas}: the outpost median likelihood under a core-trained
model is exactly zero, driven by unseen regional events in 53.1\% of
outpost strings, while held-out core strings score $1.85\times10^{-8}$.
Regional emblem practice alone generates the same form of separation once
read as dialectal adaptation of a language.

Appendix~\ref{app:methods} presents the full curves, stored null draws,
and sensitivity cells behind these summaries, whose agreement is
mechanical: the same urns, slots, licenses, and conventions surface in
each statistic from a different angle.

\subsection{Dictionary coverage does not identify a reading}

The multilingual stress test targets decipherment's evidential move
directly. Under the literal protocol, in-sample coverage
reaches 83.0\% of texts for English, 88.13\% for Sanskrit, and 67.73\% for
Tamil; grouped-test coverage falls to 74.23\%, 61.01\%, and 44.05\%, while
the random-key means are 6.50\%, 5.77\%, and 9.48\%. The three languages
are mutually incompatible readings of one corpus, yet the same search
finds substantial dictionary-supported coverage in each, on signs that
encode nothing.

The Sanskrit script and adaptive protocols push in-sample coverage to
97.10\% and 97.33\%, with grouped-test values of 93.17\% and 93.39\%. That
persistence reflects the permissiveness of the declared script and
morphological operations rather than a recovered language: random script
keys already average 27.87\%, and the fixed-key ablation of
Appendix~\ref{app:decoder-diagnostics} attributes most of the gain to
optional inherent \emph{a} and mechanical morphology.

High coverage coexists with weak identification. Only seven English signs
and one literal Sanskrit sign have singleton candidate intersections, and
Tamil has none at all. Six bootstrap script keys agree on just 9.30\% of
sign values even though each parses more than 93.5\% of the corpus, and
for signs observed at most twice, agreement falls below the 1-in-49 chance
rate.

We report no semantic-recovery score: the dictionary definitions and
\sigil{}'s administrative roles share no predeclared ontology, so any post
hoc mapping would manufacture its own target. Appendices
\ref{app:decoder} and \ref{app:decoder-diagnostics} expose the
denominators, controls, ambiguity, and stability results in full.

\section{Discussion}\label{sec:discussion}

\paragraph{What the construction establishes.}
The reported outcomes are not sufficient conditions for language: a system
of emblems with compositional administrative meaning can occupy the same
entropy regions, display the same frequency and positional concentration,
support comparable prediction and restoration, enter the writing region of
a published classifier, and produce significant directional and
lexical-distribution results, all under one fixed seed. The registry
makes the statement auditable: every eligible source outcome is matched,
and every ineligible method remains visible with its status explained.

\paragraph{What it leaves open.}
An existence witness cannot say how common such systems are, and a feature
may yet be informative if calibrated linguistic populations produce it far
more often than a weighted family of non-languages; no study evaluated
here supplies that family, and neither do we. What survives is a
compatibility claim, a weaker thing than a diagnosis.

\paragraph{Construction and circularity.}
\sigil{} was designed with the published criteria in view, and the
canonical realization is not held out. In a predictive
benchmark that would be a defect; for a constructive counterexample it is
the entire point: were the statistics logically specific to language, no
explicit non-language could be built to satisfy them. The 27-cell sweep
shows the headline behavior persisting across the declared parameter
changes.

\paragraph{Stronger evidence for decipherment.}
None of this makes decipherment hopeless; it clarifies what persuasive
evidence looks like. A proposal gains force when it commits to sign values
and semantic predictions before testing, succeeds on held-out
inscriptions, prefers one language over matched alternatives, recovers
stable values across resamples, and predicts external archaeological or
textual facts. Computational decipherment has met such standards on
ciphers, Ugaritic, and Linear B
\citep{knight1999scripts,knight2006unsupervised,knight2011,snyder2010lost,%
bergkirkpatrick2013restarts,luo2019neural}. Dictionary coverage alone
does not approach them.

\section{Conclusion}\label{sec:conclusion}

A known non-language with explicit compositional meanings receives every
exact source-validated outcome in the 54-method registry, reproduces
a much wider descriptive profile, and can be made readable in three
unrelated languages. The measures detect organization,
predictability, concentration, direction, and model misfit; they do not,
on their own, establish encoded speech. Nothing here settles the status of
the Indus
script, but the burden of proof has moved: a claim that regularity
establishes language now needs a discriminative comparison against
structured non-linguistic alternatives, or predictions they cannot readily
satisfy. Until then, the regularities remain compatible with language and
with practices that never encoded it.

\section*{Limitations}

\sigil{} is a single synthetic system, deliberately constructed after the
relevant literature was known, and its canonical realization is not a
held-out sample. The 27 sensitivity cells share a starting seed and diverge
through parameter-dependent control flow, so they display robustness without
estimating sampling variance. The work establishes compatibility between the
tested outcomes and one defined non-language; it does not estimate how
frequently comparable emblem systems would produce the same outcomes, and it
supplies no likelihood ratio between linguistic and non-linguistic
hypotheses.

The Indus corpora behind the source studies differ in cleaning, orientation,
and inventory, and our exact scoring is restricted to methods whose public
source data and decision rule we could reproduce. The resulting denominator
holds six dependent outcomes from three method families, and the other 48
registry entries remain outside it for stated reasons. The absence of a
method from the denominator must not be read as a failed Indus claim or as a
successful \sigil{} match.

Several analyses are reconstructions because the publications omit
operational details, and although we declare our tie rules, filtering,
randomization, and denominators, another reasonable reconstruction could
produce different continuous values. The sequential lexical experiment is
especially underdetermined by its source: its beam policy, complete
segmentation requirement, held-out split, and tie rules are our explicit
choices, and its coverage figures measure parse feasibility rather than
translation accuracy.

The fictional archaeological setting is not a model of Harappan society. Its
parallels show that non-phonological administrative mechanisms are coherent,
not that such mechanisms generated the Indus corpus. Finally, the study
neither compares fully specified population distributions for linguistic and
non-linguistic hypotheses nor offers any decipherment of its own.

\bibliography{refs}

\begin{thebibliography}{67}
\providecommand{\natexlab}[1]{#1}

\bibitem[{Ameri(2013)}]{ameri2013}
Marta Ameri. 2013.
\newblock Regional diversity in the {Harappan} world: The evidence of the
  seals.
\newblock In Shinu~Anna Abraham, Praveena Gullapalli, Teresa~P. Raczek, and
  Uzma~Z. Rizvi, editors, \emph{Connections and Complexity: New Approaches to
  the Archaeology of South Asia}, pages 355--374. Left Coast Press, Walnut
  Creek, CA.

\bibitem[{Ashraf and Sinha(2012)}]{ashraf2012core}
Md~Izhar Ashraf and Sitabhra Sinha. 2012.
\newblock \href {https://doi.org/10.1007/978-3-642-28604-9_12} {Core-periphery
  organization of graphemes in written sequences: decreasing positional
  rigidity with increasing core order}.
\newblock In \emph{Computational Linguistics and Intelligent Text Processing
  (CICLing 2012), LNCS 7181}, pages 142--153.

\bibitem[{Ashraf and Sinha(2018)}]{ashraf2018handedness}
Md.~Izhar Ashraf and Sitabhra Sinha. 2018.
\newblock \href {https://doi.org/10.1371/journal.pone.0190735} {The
  ``handedness'' of language: Directional symmetry breaking of sign usage in
  words}.
\newblock \emph{PLOS ONE}, 13(1):e0190735.

\bibitem[{Berg-Kirkpatrick and Klein(2013)}]{bergkirkpatrick2013restarts}
Taylor Berg-Kirkpatrick and Dan Klein. 2013.
\newblock \href {https://aclanthology.org/D13-1087/} {Decipherment with a
  million random restarts}.
\newblock In \emph{Proceedings of the 2013 Conference on Empirical Methods in
  Natural Language Processing}, pages 874--878, Seattle, Washington, USA.

\bibitem[{Chen and Goodman(1998)}]{chen1998}
Stanley~F. Chen and Joshua Goodman. 1998.
\newblock \href {https://dash.harvard.edu/handle/1/25104739} {An empirical
  study of smoothing techniques for language modeling}.
\newblock Technical Report TR-10-98, Harvard Computer Science Group.

\bibitem[{Collon(1987)}]{collon1987}
Dominique Collon. 1987.
\newblock \emph{First Impressions: Cylinder Seals in the Ancient Near East}.
\newblock British Museum Publications, London.

\bibitem[{Dunning(1993)}]{dunning1993}
Ted Dunning. 1993.
\newblock \href {https://aclanthology.org/J93-1003/} {Accurate methods for the
  statistics of surprise and coincidence}.
\newblock \emph{Computational Linguistics}, 19(1):61--74.

\bibitem[{Englund(2011)}]{englund2011}
Robert~K. Englund. 2011.
\newblock \href {https://doi.org/10.1093/oxfordhb/9780199557301.013.0002}
  {Accounting in proto-cuneiform}.
\newblock In Karen Radner and Eleanor Robson, editors, \emph{The Oxford
  Handbook of Cuneiform Culture}, pages 32--50. Oxford University Press,
  Oxford.

\bibitem[{Farmer et~al.(2004)Farmer, Sproat, and Witzel}]{farmer2004}
Steve Farmer, Richard Sproat, and Michael Witzel. 2004.
\newblock \href {https://doi.org/10.11588/ejvs.2004.2.620} {The collapse of the
  {Indus}-script thesis: the myth of a literate {Harappan} civilization}.
\newblock \emph{Electronic Journal of Vedic Studies}, 11(2):19--57.

\bibitem[{Francis and Ku{\v c}era(1979)}]{francis1979}
W.~Nelson Francis and Henry Ku{\v c}era. 1979.
\newblock \emph{Brown Corpus Manual: Manual of Information to Accompany a
  Standard Corpus of Present-Day Edited American English, for Use with Digital
  Computers}.
\newblock Department of Linguistics, Brown University, Providence, RI.

\bibitem[{Fuls(2015)}]{fuls2015}
Andreas Fuls. 2015.
\newblock \href {https://doi.org/10.13109/hisp.2015.128.1.42} {Classifying
  undeciphered writing systems}.
\newblock \emph{Historische Sprachforschung}, 128(1):42--58.

\bibitem[{Gelb(1952)}]{gelb1952}
Ignace~J. Gelb. 1952.
\newblock \emph{A Study of Writing: The Foundations of Grammatology}.
\newblock University of Chicago Press, Chicago.

\bibitem[{Glatz(2012)}]{glatz2012}
Claudia Glatz. 2012.
\newblock \href {https://doi.org/10.3764/aja.116.1.0005} {Bearing the marks of
  control? reassessing pot marks in late bronze age {Anatolia}}.
\newblock \emph{American Journal of Archaeology}, 116(1):5--38.

\bibitem[{Gordon(2026)}]{gordon2026}
Michael Gordon. 2026.
\newblock Cracking the {Indus} script.
\newblock Popular Archaeology, Summer 2026 issue,
  \url{https://popular-archaeology.com/article/cracking-the-indus-script-2/}.

\bibitem[{Khanna and Merriam(2025)}]{khanna2025}
Ruhan Khanna and Louie Merriam. 2025.
\newblock \href {https://doi.org/10.5120/ijca2025926075} {A computational
  analysis of the {Indus} script: Identifying sign functions in logo-syllabic
  writing systems}.
\newblock \emph{International Journal of Computer Applications},
  187(64):19--22.

\bibitem[{Kneser and Ney(1995)}]{kneser1995}
Reinhard Kneser and Hermann Ney. 1995.
\newblock \href {https://doi.org/10.1109/ICASSP.1995.479394} {Improved
  backing-off for $m$-gram language modeling}.
\newblock In \emph{Proceedings of the 1995 International Conference on
  Acoustics, Speech, and Signal Processing ({ICASSP})}, volume~1, pages
  181--184. IEEE.

\bibitem[{Knight et~al.(2011)Knight, Megyesi, and Schaefer}]{knight2011}
Kevin Knight, Be{\'a}ta Megyesi, and Christiane Schaefer. 2011.
\newblock \href {https://aclanthology.org/W11-1202/} {The {Copiale} cipher}.
\newblock In \emph{Proceedings of the 4th Workshop on Building and Using
  Comparable Corpora: Comparable Corpora and the Web}, pages 2--9, Portland,
  Oregon. Association for Computational Linguistics.

\bibitem[{Knight et~al.(2006)Knight, Nair, Rathod, and
  Yamada}]{knight2006unsupervised}
Kevin Knight, Anish Nair, Nishit Rathod, and Kenji Yamada. 2006.
\newblock \href {https://aclanthology.org/P06-2065/} {Unsupervised analysis for
  decipherment problems}.
\newblock In \emph{Proceedings of the {COLING}/{ACL} 2006 Main Conference
  Poster Sessions}, pages 499--506, Sydney, Australia.

\bibitem[{Knight and Yamada(1999)}]{knight1999scripts}
Kevin Knight and Kenji Yamada. 1999.
\newblock \href {https://aclanthology.org/W99-0906/} {A computational approach
  to deciphering unknown scripts}.
\newblock In \emph{Unsupervised Learning in Natural Language Processing}.

\bibitem[{Koskenniemi(1981)}]{koskenniemi1981}
Kimmo Koskenniemi. 1981.
\newblock \href {https://journal.fi/store/article/view/49897} {Syntactic
  methods in the study of the {Indus} script}.
\newblock In \emph{Proceedings of the Nordic South Asia Conference Held in
  Helsinki, 10--12 June 1980}, volume~50 of \emph{Studia Orientalia}, pages
  125--136, Helsinki. Finnish Oriental Society.

\bibitem[{Laursen(2010)}]{laursen2010}
Steffen~Terp Laursen. 2010.
\newblock \href {https://doi.org/10.1111/j.1600-0471.2010.00329.x} {The
  westward transmission of {Indus Valley} sealing technology: Origin and
  development of the `{Gulf Type}' seal and other administrative technologies
  in early {Dilmun}, c.~2100--2000 {BC}}.
\newblock \emph{Arabian Archaeology and Epigraphy}, 21(2):96--134.

\bibitem[{Lee et~al.(2010{\natexlab{a}})Lee, Jonathan, and
  Ziman}]{lee2010pictish}
Rob Lee, Philip Jonathan, and Pauline Ziman. 2010{\natexlab{a}}.
\newblock \href {https://doi.org/10.1098/rspa.2010.0041} {Pictish symbols
  revealed as a written language through application of {Shannon} entropy}.
\newblock \emph{Proceedings of the Royal Society A: Mathematical, Physical and
  Engineering Sciences}, 466(2121):2545--2560.

\bibitem[{Lee et~al.(2010{\natexlab{b}})Lee, Jonathan, and Ziman}]{lee2010cl}
Rob Lee, Philip Jonathan, and Pauline Ziman. 2010{\natexlab{b}}.
\newblock \href {https://doi.org/10.1162/coli_c_00029} {A response to {Richard
  Sproat} on random systems, writing, and entropy}.
\newblock \emph{Computational Linguistics}, 36(4):791--794.

\bibitem[{Liberman(2009)}]{liberman2009}
Mark Liberman. 2009.
\newblock Conditional entropy and the {Indus} script.
\newblock Language Log, \url{https://languagelog.ldc.upenn.edu/nll/?p=1374}.

\bibitem[{Liberman(2010)}]{liberman2010}
Mark Liberman. 2010.
\newblock Pictish writing?
\newblock Language Log, \url{https://languagelog.ldc.upenn.edu/nll/?p=2227}.

\bibitem[{Luo et~al.(2019)Luo, Cao, and Barzilay}]{luo2019neural}
Jiaming Luo, Yuan Cao, and Regina Barzilay. 2019.
\newblock \href {https://doi.org/10.18653/v1/P19-1303} {Neural decipherment via
  minimum-cost flow: From {U}garitic to {L}inear {B}}.
\newblock In \emph{Proceedings of the 57th Annual Meeting of the Association
  for Computational Linguistics}, pages 3146--3155, Florence, Italy.

\bibitem[{Mahadevan(1977)}]{mahadevan1977}
Iravatham Mahadevan. 1977.
\newblock \emph{The {Indus} Script: Texts, Concordance and Tables}.
\newblock Number~77 in Memoirs of the Archaeological Survey of India.
  Archaeological Survey of India, New Delhi.

\bibitem[{Mandelbrot(1953)}]{mandelbrot1953}
Beno{\^\i}t Mandelbrot. 1953.
\newblock An informational theory of the statistical structure of language.
\newblock In Willis Jackson, editor, \emph{Communication Theory}, pages
  486--502. Butterworths Scientific Publications, London.

\bibitem[{Monier-Williams(1899)}]{monier1899}
Monier Monier-Williams. 1899.
\newblock \emph{A {Sanskrit--English} Dictionary: Etymologically and
  Philologically Arranged with Special Reference to Cognate Indo-European
  Languages}.
\newblock Clarendon Press, Oxford.
\newblock Cologne Digital Sanskrit Dictionaries digitization,
  \url{https://www.sanskrit-lexicon.uni-koeln.de/}.

\bibitem[{Mukhopadhyay(2019)}]{mukhopadhyay2019}
Bahata~Ansumali Mukhopadhyay. 2019.
\newblock \href {https://doi.org/10.1057/s41599-019-0274-1} {Interrogating
  {Indus} inscriptions to unravel their mechanisms of meaning conveyance}.
\newblock \emph{Palgrave Communications}, 5(1):73.

\bibitem[{Mukul(2025)}]{indiatoday2025}
Sushim Mukul. 2025.
\newblock \href
  {https://www.indiatoday.in/india/tamil-nadu/story/tamil-nadu-cm-mk-stalin-usd-1-million-decode-decipher-indus-valley-civilisation-harappan-script-explained-why-2660891-2025-01-07}
  {Why {Tamil Nadu} {CM} {Stalin} pledged {USD} 1 million for decoding {Indus}
  script}.
\newblock India Today, 7 January 2025.

\bibitem[{Nair(2026)}]{nair2026}
Ashish Nair. 2026.
\newblock \href {https://arxiv.org/abs/2604.17828} {How non-linguistic is the
  {Indus} sign system? a synthetic-baseline scorecard}.
\newblock arXiv preprint arXiv:2604.17828.
\newblock \emph{Preprint}, arXiv:2604.17828.

\bibitem[{Nemenman et~al.(2002)Nemenman, Shafee, and Bialek}]{nsb2002}
Ilya Nemenman, Fariel Shafee, and William Bialek. 2002.
\newblock \href {https://doi.org/10.7551/mitpress/1120.003.0065} {Entropy and
  inference, revisited}.
\newblock In \emph{Advances in Neural Information Processing Systems 14}, pages
  471--478. MIT Press.

\bibitem[{Oakes(2019)}]{oakes2019}
Michael~P. Oakes. 2019.
\newblock \href {https://doi.org/10.1080/09296174.2017.1406294} {Statistical
  analysis of the tables in {Mahadevan}'s concordance of the {Indus Valley
  Script}}.
\newblock \emph{Journal of Quantitative Linguistics}, 26(1):22--47.

\bibitem[{Paninski(2003)}]{paninski2003}
Liam Paninski. 2003.
\newblock \href {https://doi.org/10.1162/089976603321780272} {Estimation of
  entropy and mutual information}.
\newblock \emph{Neural Computation}, 15(6):1191--1253.

\bibitem[{Parpola(1994)}]{parpola1994}
Asko Parpola. 1994.
\newblock \emph{Deciphering the {Indus} Script}.
\newblock Cambridge University Press, Cambridge.

\bibitem[{Pitman and Yor(1997)}]{pitman1997}
Jim Pitman and Marc Yor. 1997.
\newblock \href {https://doi.org/10.1214/aop/1024404422} {The two-parameter
  {Poisson--Dirichlet} distribution derived from a stable subordinator}.
\newblock \emph{The Annals of Probability}, 25(2):855--900.

\bibitem[{Rajan(2024)}]{aksharamukha}
Vinodh Rajan. 2024.
\newblock Aksharamukha: script converter.
\newblock \url{https://www.aksharamukha.com/}.
\newblock Version 2.3.

\bibitem[{Ramani and Kishor(2024)}]{ramani2024}
R.~Geetha Ramani and S.~Kishor. 2024.
\newblock \href {https://doi.org/10.5120/ijca2024923836} {Deciphering the
  {Indus Valley} script: Hierarchical clustering and dependency tree analysis}.
\newblock \emph{International Journal of Computer Applications}, 186(31):5--16.

\bibitem[{Rao(2010)}]{rao2010ieee}
Rajesh P.~N. Rao. 2010.
\newblock \href {https://doi.org/10.1109/MC.2010.112} {Probabilistic analysis
  of an ancient undeciphered script}.
\newblock \emph{Computer}, 43(4):76--80.

\bibitem[{Rao et~al.(2015)Rao, Lee, Yadav, Vahia, Jonathan, and
  Ziman}]{rao2015language}
Rajesh P.~N. Rao, Rob Lee, Nisha Yadav, Mayank~N. Vahia, Philip Jonathan, and
  Pauline Ziman. 2015.
\newblock \href {https://doi.org/10.1353/lan.2015.0055} {On statistical
  measures and ancient writing systems}.
\newblock \emph{Language}, 91(4):e198--e205.

\bibitem[{Rao et~al.(2009{\natexlab{a}})Rao, Yadav, Vahia, Joglekar, Adhikari,
  and Mahadevan}]{rao2009science}
Rajesh P.~N. Rao, Nisha Yadav, Mayank~N. Vahia, Hrishikesh Joglekar, Ronojoy
  Adhikari, and Iravatham Mahadevan. 2009{\natexlab{a}}.
\newblock \href {https://doi.org/10.1126/science.1170391} {Entropic evidence
  for linguistic structure in the {Indus} script}.
\newblock \emph{Science}, 324(5931):1165.

\bibitem[{Rao et~al.(2009{\natexlab{b}})Rao, Yadav, Vahia, Joglekar, Adhikari,
  and Mahadevan}]{rao2009pnas}
Rajesh P.~N. Rao, Nisha Yadav, Mayank~N. Vahia, Hrishikesh Joglekar, Ronojoy
  Adhikari, and Iravatham Mahadevan. 2009{\natexlab{b}}.
\newblock \href {https://doi.org/10.1073/pnas.0906237106} {A {Markov} model of
  the {Indus} script}.
\newblock \emph{Proceedings of the National Academy of Sciences},
  106(33):13685--13690.

\bibitem[{Rao et~al.(2010)Rao, Yadav, Vahia, Joglekar, Adhikari, and
  Mahadevan}]{rao2010cl}
Rajesh P.~N. Rao, Nisha Yadav, Mayank~N. Vahia, Hrishikesh Joglekar, Ronojoy
  Adhikari, and Iravatham Mahadevan. 2010.
\newblock \href {https://doi.org/10.1162/coli_c_00030} {Entropy, the {Indus}
  script, and language: a reply to {R. Sproat}}.
\newblock \emph{Computational Linguistics}, 36(4):795--805.

\bibitem[{Recchia and Louwerse(2016)}]{recchia2016}
Gabriel~L. Recchia and Max~M. Louwerse. 2016.
\newblock \href {https://doi.org/10.1111/cogs.12311} {Archaeology through
  computational linguistics: Inscription statistics predict excavation sites of
  {Indus Valley} artifacts}.
\newblock \emph{Cognitive Science}, 40(8):2065--2080.

\bibitem[{Shannon(1949)}]{shannon1949}
Claude~E. Shannon. 1949.
\newblock \href {https://doi.org/10.1002/j.1538-7305.1949.tb00928.x}
  {Communication theory of secrecy systems}.
\newblock \emph{Bell System Technical Journal}, 28(4):656--715.

\bibitem[{Shannon(1951)}]{shannon1951}
Claude~E. Shannon. 1951.
\newblock \href {https://doi.org/10.1002/j.1538-7305.1951.tb01366.x}
  {Prediction and entropy of printed {English}}.
\newblock \emph{Bell System Technical Journal}, 30(1):50--64.

\bibitem[{Sinha et~al.(2011)Sinha, Ashraf, Pan, and Wells}]{sinha2011network}
Sitabhra Sinha, Md~Izhar Ashraf, Raj~Kumar Pan, and Bryan~K. Wells. 2011.
\newblock \href {https://doi.org/10.1016/j.csl.2010.05.007} {Network analysis
  of a corpus of undeciphered {Indus} civilization inscriptions indicates
  syntactic organization}.
\newblock \emph{Computer Speech \& Language}, 25(3):639--654.

\bibitem[{Siromoney and Huq(1988)}]{siromoney1988}
Gift Siromoney and Abdul Huq. 1988.
\newblock \href {https://doi.org/10.1007/BF00056346} {Segmentation of {Indus}
  texts: A dynamic programming approach}.
\newblock \emph{Computers and the Humanities}, 22(1):11--21.

\bibitem[{Snyder et~al.(2010)Snyder, Barzilay, and Knight}]{snyder2010lost}
Benjamin Snyder, Regina Barzilay, and Kevin Knight. 2010.
\newblock \href {https://aclanthology.org/P10-1107/} {A statistical model for
  lost language decipherment}.
\newblock In \emph{Proceedings of the 48th Annual Meeting of the Association
  for Computational Linguistics}, pages 1048--1057, Uppsala, Sweden.

\bibitem[{Speer(2022)}]{speer2022wordfreq}
Robyn Speer. 2022.
\newblock \href {https://doi.org/10.5281/zenodo.7199437} {rspeer/wordfreq:
  v3.0}.
\newblock Zenodo.

\bibitem[{Sproat(2010{\natexlab{a}})}]{sproat2010cl}
Richard Sproat. 2010{\natexlab{a}}.
\newblock \href {https://doi.org/10.1162/coli_a_00011} {Ancient symbols,
  computational linguistics, and the reviewing practices of the general science
  journals}.
\newblock \emph{Computational Linguistics}, 36(3):585--594.

\bibitem[{Sproat(2010{\natexlab{b}})}]{sproat2010reply}
Richard Sproat. 2010{\natexlab{b}}.
\newblock \href {https://doi.org/10.1162/coli_c_00031} {Reply to {Rao} et al.\
  and {Lee} et al.}
\newblock \emph{Computational Linguistics}, 36(4):807--816.

\bibitem[{Sproat(2014)}]{sproat2014language}
Richard Sproat. 2014.
\newblock \href {https://doi.org/10.1353/lan.2014.0031} {A statistical
  comparison of written language and nonlinguistic symbol systems}.
\newblock \emph{Language}, 90(2):457--481.

\bibitem[{Sproat(2015)}]{sproat2015reply}
Richard Sproat. 2015.
\newblock \href {https://doi.org/10.1353/lan.2015.0058} {On misunderstandings
  and misrepresentations: a reply to {Rao} et al.}
\newblock \emph{Language}, 91(4):e206--e208.

\bibitem[{Tiwari(2026)}]{tiwari2026}
Tanishk Tiwari. 2026.
\newblock \href {https://doi.org/10.18653/v1/2026.nlp4dh-1.28} {Statistical
  structure in {Indus} sign sequences}.
\newblock In \emph{Proceedings of the 6th International Conference on Natural
  Language Processing for the Digital Humanities}, pages 314--319. Association
  for Computational Linguistics.

\bibitem[{Venkatesh and Farghaly(2023{\natexlab{a}})}]{venkatesh2023westasia}
Varun Venkatesh and Ali Farghaly. 2023{\natexlab{a}}.
\newblock \href {https://doi.org/10.1109/ICCCNT56998.2023.10306531}
  {Identifying anomalous {Indus} texts from {West Asia} using {Markov} chain
  language models}.
\newblock In \emph{2023 14th International Conference on Computing
  Communication and Networking Technologies ({ICCCNT})}, pages 1--7.

\bibitem[{Venkatesh and Farghaly(2023{\natexlab{b}})}]{venkatesh2023}
Varun Venkatesh and Ali Farghaly. 2023{\natexlab{b}}.
\newblock \href {https://doi.org/10.59720/22-256} {Statistical models for
  identifying missing and unclear signs of the {Indus} script}.
\newblock \emph{Journal of Emerging Investigators}, 6.

\bibitem[{Vidale(2007)}]{vidale2007}
Massimo Vidale. 2007.
\newblock The collapse melts down: a reply to {Farmer, Sproat and Witzel}.
\newblock \emph{East and West}, 57(1--4):333--366.

\bibitem[{Wells and Fuls(2006)}]{wells2015icit}
Bryan~K. Wells and Andreas Fuls. 2006.
\newblock The interactive corpus of {Indus} text ({ICIT}).
\newblock Online corpus, \url{https://www.epigraphica.de/indus/menueindus.htm}.

\bibitem[{Yadav et~al.(2010)Yadav, Joglekar, Rao, Vahia, Adhikari, and
  Mahadevan}]{yadav2010plos}
Nisha Yadav, Hrishikesh Joglekar, Rajesh P.~N. Rao, Mayank~N. Vahia, Ronojoy
  Adhikari, and Iravatham Mahadevan. 2010.
\newblock \href {https://doi.org/10.1371/journal.pone.0009506} {Statistical
  analysis of the {Indus} script using $n$-grams}.
\newblock \emph{PLoS ONE}, 5(3):e9506.

\bibitem[{Yadav et~al.(2017)Yadav, Salgaonkar, and Vahia}]{yadav2017clusters}
Nisha Yadav, Ambuja Salgaonkar, and Mayank Vahia. 2017.
\newblock \href {https://doi.org/10.5120/ijca2017913207} {Clustering {Indus}
  texts using k-means}.
\newblock \emph{International Journal of Computer Applications}, 162(1):16--21.

\bibitem[{Yadav et~al.(2008{\natexlab{a}})Yadav, Vahia, Mahadevan, and
  Joglekar}]{yadav2008segmentation}
Nisha Yadav, Mayank~N. Vahia, Iravatham Mahadevan, and Hrishikesh Joglekar.
  2008{\natexlab{a}}.
\newblock Segmentation of {Indus} texts.
\newblock \emph{International Journal of Dravidian Linguistics}, 37(1):53--72.

\bibitem[{Yadav et~al.(2008{\natexlab{b}})Yadav, Vahia, Mahadevan, and
  Joglekar}]{yadav2008pattern}
Nisha Yadav, Mayank~N. Vahia, Iravatham Mahadevan, and Hrishikesh Joglekar.
  2008{\natexlab{b}}.
\newblock A statistical approach for pattern search in {Indus} writing.
\newblock \emph{International Journal of Dravidian Linguistics}, 37(1):39--52.

\bibitem[{{Yajnadevam}(2024)}]{yajnadevam2024}
{Yajnadevam}. 2024.
\newblock A cryptanalytic decipherment of the {Indus} script.
\newblock Self-published preprint, version compiled 13 November 2024;
  \url{https://rarebooksocietyofindia.org/book_archive/A_cryptanalytic_decipherment_of_the_Indu.pdf}.
\newblock Released code: \url{https://github.com/yajnadevam/ScriptDerivation},
  revision 5835e8b.

\bibitem[{Zettler(1987)}]{zettler1987}
Richard~L. Zettler. 1987.
\newblock \href {https://doi.org/10.2307/1359781} {Sealings as artifacts of
  institutional administration in ancient {Mesopotamia}}.
\newblock \emph{Journal of Cuneiform Studies}, 39(2):197--240.

\bibitem[{Zipf(1949)}]{zipf1949}
George~Kingsley Zipf. 1949.
\newblock \emph{Human Behavior and the Principle of Least Effort: An
  Introduction to Human Ecology}.
\newblock Addison-Wesley Press, Cambridge, MA.

\end{thebibliography}

\appendix

\section{Generator and Corpus Contract}\label{app:generator}

The public corpus artifact stores all 3,000 core and 350 outpost records.
Every record carries five aligned fields: the semantic reading order, the
physical surface order, a normalized right-to-left spatial order, artifact
metadata, and the compositional semantics. The registry, ligature
definitions, weighted epithet licenses, regional patron identities, issue
quotas, urn histories, and generator settings are stored in the same
deterministic compressed JSON artifact, so every analysis in this paper can
be traced from its inputs. The local analysis cache is a separate file and
never enters the public archive.

The composition stream first creates 1,200 core texts. Every adjacent pair
observed at least 25 times in that phase becomes eligible for a registered
ligature, and later texts write an eligible pair as a fused sign with
probability 0.25. This rule discovers exactly 17 ligatures. A ligature
replaces two written emblems while preserving both underlying semantic
roles, and deity doubling likewise remains an explicit emphasis operation
in the semantic record even when one member of the doubled pair is written
inside a ligature.

For each dynamic category, a finite-pool Pitman-Yor urn introduces new types
and reinforces old ones. When a category's type shelf is exhausted, a
proposed new-type event is redirected to the existing-type distribution;
this finite-pool rule is part of the declared model rather than an
implementation fallback. Fixed weighted choices govern openers, ranks,
tallies, and terminals, and the four core regions carry weights of 0.38,
0.27, 0.20, and 0.15, realizing 1,122, 808, 606, and 464 texts. Weighted
epithet licenses are drawn with replacement during license construction, so
a license may contain repeated epithet identities. Of the 241 licenses, 105
hold a single epithet, 65 hold two, 26 three, 18 four, and 27 five, and 48
contain at least one repeated entry, which is why particular deity and
epithet pairs recur so strongly.

The outpost draws its 350 texts from the same generator under looser
conventions. Its reinforced support urn realizes 143 seal matrices, 129
clay sealings, and 78 vessel inscriptions, its mean text length is 4.363
signs against the core's 4.572, and its 335 distinct strings ground the
regional likelihood experiment of Appendix~\ref{app:methods}. No
administrative issue mark is ever assigned there, since issue quotas bind
core clay sealings and vessels alone.

Administrative issue signs use their own seed. Each of five issue rounds in
each of four core regions receives a quota of $4+\mathrm{Binomial}(8,0.5)$
uses, assigned without replacement among that region's clay sealings and
vessel inscriptions, for a realized total of 157 uses across the corpus.
Each mark is placed immediately before the last primary sign of its text.
The mechanism supplies a declared administrative reason for rare,
positionally restricted types, which several Indus analyses would otherwise
treat as evidence about phonology.

\begin{table}[H]
\centering
\small
\begin{tabular}{@{}lll@{}}
\toprule
Property & \sigil{} & EBUDS \\
\midrule
Texts & 3,000 & 1,548 \\
Tokens & 13,717 & about 7,000 \\
Attested / nominal types & 419 / 450 & 377 / 417 \\
Mean length & 4.572 & about 4.5 \\
Top-sign share & 6.99\% & about 10\% \\
Signs covering 80\% & 51 & 69 \\
Initial signs covering 80\% & 39 & 82 \\
Final signs covering 80\% & 10 & 23 \\
Texts with a repeated sign & 13.0\% & about 17\% \\
\bottomrule
\end{tabular}
\caption{Corpus-level descriptive quantities. The EBUDS values are the
anchors reported in the cited statistical literature
\citep{rao2009pnas,yadav2010plos}. The comparison is descriptive, not an
exact score.}
\label{tab:corpus-appendix}
\end{table}

Table~\ref{tab:corpus-appendix} places the two corpora side by side. The
canonical corpus is larger than EBUDS but stays within the same regime of
short texts and a few hundred sign types, which is the regime in which the
published statistics operate. Inventory size and text length are design
inputs; the rank-frequency shape, the positional concentration, and the
remaining results arise from the realized composition within that regime.

Two small examples convey what the aligned records contain. Core text 1
consists of three signs read as an auspicious opener, the deity Hare-Dawn,
and a validating seal knot, and its semantic record stores exactly that
dedication. Another core text writes the deity Tortoise-Flood, the potter's
guild emblem, the grain commodity sign, a one-unit tally, and a terminal,
and its record stores the corresponding entitlement of one measure of grain
to a potter under that patron. The stored semantics are what the decoder of
Appendix~\ref{app:decoder} never sees.

\section{Statistical Methods and Extended Results}\label{app:methods}

\paragraph{Conditional entropy.}
For $k=20,40,\ldots,400$, each corpus retains its $k$ most frequent tokens
and fits an interpolated modified Kneser-Ney bigram estimator. Filtering
preserves gaps, so tokens separated by an excluded item do not become
adjacent, and the effective inventory after filtering is reported at every
point, which is why the plotted curves end at different horizontal
positions. Sparse count-of-count configurations fall back to a documented
discount. The sweep follows the tradition of vocabulary-growth entropy
estimates that began with printed English \citep{shannon1951}.

\paragraph{Block entropy.}
Blocks of orders one through six are counted inside texts, never across text
boundaries. The NSB estimator integrates over its concentration parameter
with stable evidence evaluation and convergence checks, a choice motivated
by the strong finite-sample bias of naive entropy estimates on short
sequences \citep{nsb2002,paninski2003}. Values are reported in units of
$\log L$, with $L$ the alphabet size, following the reconstructed source
comparison.

\paragraph{Prediction and restoration.}
Witten-Bell $n$-gram models use a fixed vocabulary and recursive backoff on
one seeded five-fold split of distinct strings. Restoration uses a separate
seeded five-fold split in which one position is removed from each validation
text 100 times, giving 275,300 evaluated trials with no out-of-vocabulary
failures. We report top-one accuracy and membership in the smallest
candidate set whose posterior mass reaches 90\%.

\paragraph{Frequency, position, and segmentation.}
The fitted curve is $\log f_r=a-b\log(r+c)$. Positional concentration counts
the smallest sign set covering 80\% of initial or final tokens. Staged
segmentation uses distinct texts of length at least five together with the
source cutoffs for complete texts and frequent combinations. The publication
does not reproducibly specify its singleton-frequency cutoff, so we retain
both the pre-singleton result and the full sensitivity range rather than
choosing a favorable endpoint.

\paragraph{Networks and regional likelihood.}
The transition graph includes self-loops. Reciprocity and connectance use
100 stored within-text permutations under a pinned seed. The outpost analogue
trains a modified Kneser-Ney bigram model on 2,653 of the 2,753 distinct
core strings, then scores the 100 held-out core strings and all 335 distinct
outpost strings, keeping overall medians, nonzero medians, and zero shares
separately. The
held-out core strings calibrate what a text earns for being
unseen, so the outpost comparison isolates regional novelty rather than
absence from training.

\begin{figure}[!htb]
\centering
\includegraphics[width=.88\columnwidth]{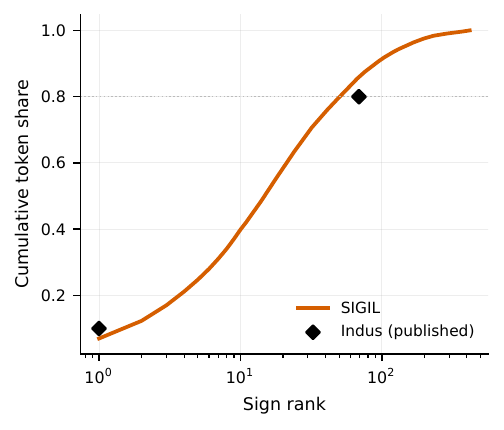}
\caption{Cumulative token coverage by sign rank for \sigil{}. The two black
diamonds are published Indus anchors: a top-sign share near 10\% and 69
signs required for 80\% coverage.}
\label{fig:coverage}
\end{figure}

\paragraph{Sign coverage (Figure~\ref{fig:coverage}).}
The most frequent \sigil{} sign contributes 6.99\% of all tokens, and 51
signs cover 80\% of them. Shuffling leaves the curve untouched because the
statistic is purely unigram-based. Frequency concentration is real corpus
structure, and it exists independently of sequence, syntax, or phonology.

\begin{figure}[!htb]
\centering
\includegraphics[width=.82\columnwidth]{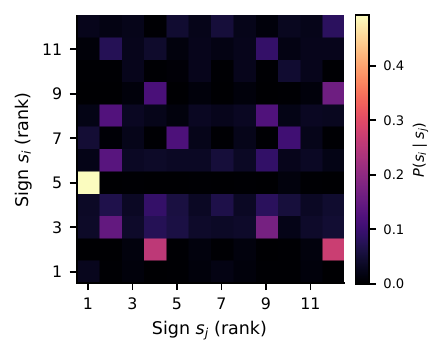}
\caption{Kneser-Ney transition probabilities among the twelve most frequent
\sigil{} signs. The largest cell, at 0.49, is the epithet Radiant after
the deity Hare-Dawn; slots and licenses leave a few preferred successors
against a sparse background.}
\label{fig:transition}
\end{figure}

\paragraph{Transitions (Figure~\ref{fig:transition}).}
A handful of high-probability successors stands against a sparse background,
which is the visual form of the local regularity that Indus analyses have
described as grammar-like \citep{rao2009pnas}. The panel's cast explains
its texture: the twelve most frequent signs comprise one deity, four
seal terminals, two tallies, two epithets, two commodities, and a guild,
and the four terminal columns carry only the smoother's background,
since a terminal never takes a successor. In \sigil{} every such
preference has a named constructional source, from epithet licensing to
tally attachment, and none of them involves a phonological value.

\begin{figure}[!htb]
\centering
\includegraphics[width=.82\columnwidth]{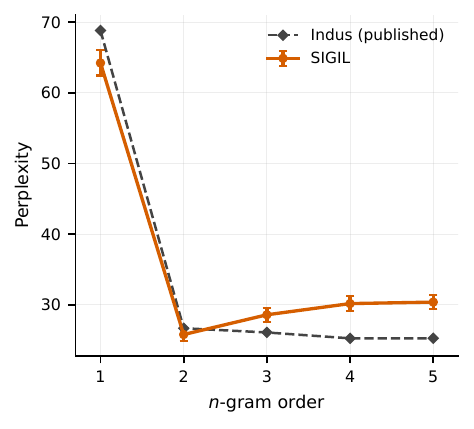}
\caption{Witten-Bell perplexity under five-fold distinct-text evaluation,
with fold variability as error bars. The large unigram-to-bigram drop
reported for the Indus corpus also occurs in \sigil{}; unlike the published
Indus values, the \sigil{} curve rises again beyond order two.}
\label{fig:perplexity}
\end{figure}

\paragraph{Perplexity (Figure~\ref{fig:perplexity}).}
Mean values by order are 64.223, 25.797, 28.592, 30.177, and 30.387, with
fold standard deviations between 0.86 and 1.77. The bigram gain faithfully
captures strong local constraint, while higher orders find too few reusable
contexts in short texts to improve the held-out score, so the curve turns
gently upward where the published Indus sequence declines further and then
flattens.

\begin{figure}[!htb]
\centering
\includegraphics[width=.82\columnwidth]{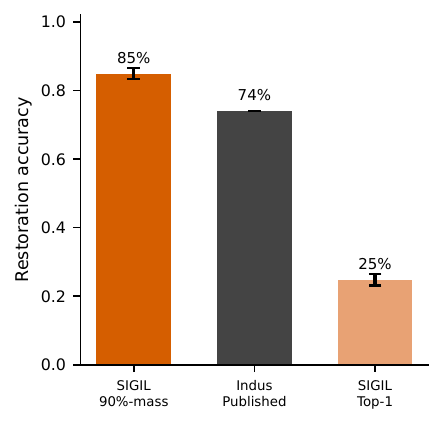}
\caption{Masked-sign restoration over 275,300 trials. The 90\%-mass
candidate criterion reaches 84.88\%, rank-one accuracy reaches 24.84\%, and
the published Indus result for the 90\%-mass criterion is 74\%.}
\label{fig:restoration}
\end{figure}

\paragraph{Restoration (Figure~\ref{fig:restoration}).}
The distance between candidate-set accuracy and rank-one accuracy is part of
the finding rather than a nuisance. A broad candidate set frequently
contains the deleted sign even when the model rarely ranks it first, so the
two scores describe different amounts of predictability, and only the
broader one approaches the published headline number.

\begin{figure}[!htb]
\centering
\includegraphics[width=\columnwidth]{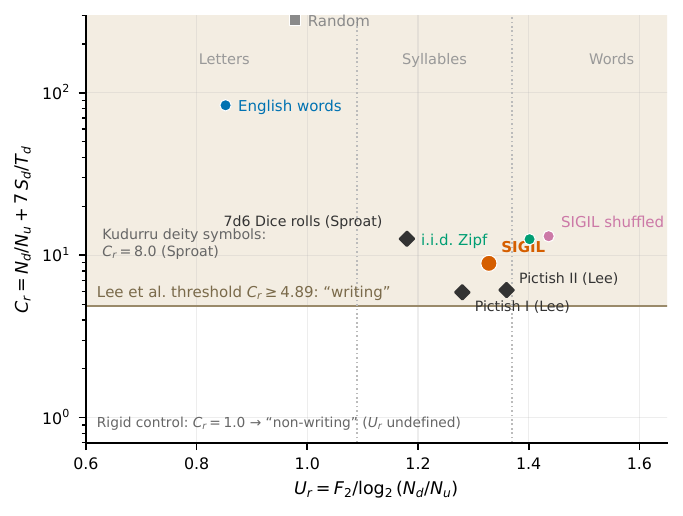}
\caption{The Lee decision plane. \sigil{} enters ``writing: syllables'';
shuffled \sigil{} is also labeled writing, in the word subclass; the rigid
control alone is non-writing, with $C_r=1$ and $U_r$ undefined. Published
Pictish outcomes and earlier counterexamples appear for context.}
\label{fig:lee}
\end{figure}

\paragraph{Classifier output (Figure~\ref{fig:lee}).}
The rule first requires $C_r\geq4.89$ and then reads the subclass from
$U_r$. \sigil{} has $C_r=8.926$ and $U_r=1.328$, which places it above
Sproat's kudurru symbols at $C_r=8.0$ and inside the syllabic band. The
published anchors calibrate that placement: the two Pictish samples
behind the original claim sit lower, at $C_r$ of 5.92 and 6.11, and
Sproat's dice-roll sequences reach $C_r=12.64$ in the same syllabic
band, so \sigil{} lies deeper inside the writing region than the corpora
the plane was drawn to classify. The
shuffled corpus passes as well, because the features preserve substantial
unigram and length information, and contemporary critiques of the Pictish
application anticipated this behavior \citep{liberman2010}. When a
classifier labels an explicit non-language as writing, it demonstrates
non-specificity for this comparison class about as directly as anything
could.

\paragraph{What the network null preserves.}
The two graph contrasts that follow share one null operation, and what
that operation holds fixed determines what the contrasts can mean. An edge
joins two
signs adjacent in semantic reading order, weighted by the number of such
adjacencies, and the graph retains its 51 self-loops. Each null draw
independently permutes the signs inside every text, which preserves text
lengths, the full unigram inventory, within-text repetition, and
co-occurrence at the text level while destroying adjacency and direction.
A large departure therefore locates order inside already fixed text-level
content, and nothing more; the test never compares the observed graph with
other ordered emblem practices, which is the comparison a linguistic
conclusion would require.

Reciprocity and connectance are also two summaries of that single
operation rather than independent probes. Slot restrictions suppress many
possible pairs while favoring a limited set of directed transitions, and
shuffling spreads the same tokens across more neighboring pairs while
creating reverse edges, so both null values rise together. Treating the
two contrasts as separate confirmations would double-count the information
in the graph.

\begin{figure}[!htb]
\centering
\includegraphics[width=.90\columnwidth]{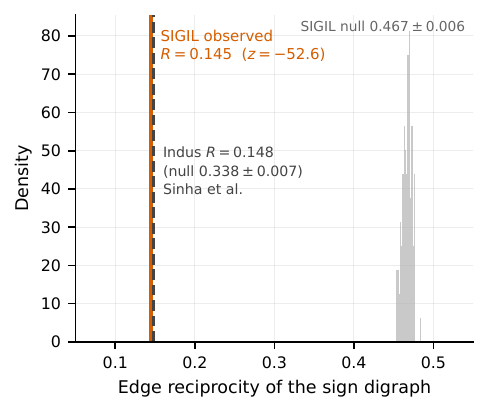}
\caption{Observed transition-graph reciprocity against 100 stored
within-text shuffles. The \sigil{} and published Indus observations both
fall far below their respective shuffle distributions.}
\label{fig:reciprocity}
\end{figure}

\paragraph{Reciprocity (Figure~\ref{fig:reciprocity}).}
\sigil{} reciprocity is 0.1449 against a null mean of 0.4669 with standard
deviation 0.0061, a departure of $z=-52.57$. Directional slots and licensed
successors produce the contrast. The published Indus comparison has the
same shape: 0.148 observed against a shuffle mean of
0.338 with standard deviation 0.007, roughly 27 null standard deviations
below its own null, and the figure plots both observations beside their
stored distributions. Deity signs precede their licensed
epithets and commodities precede their tallies, so most ordered pairs
have no returning edge. The test is strong evidence against this
particular within-text permutation and weak evidence about what
generated the order it detects.

\begin{figure}[!htb]
\centering
\includegraphics[width=.90\columnwidth]{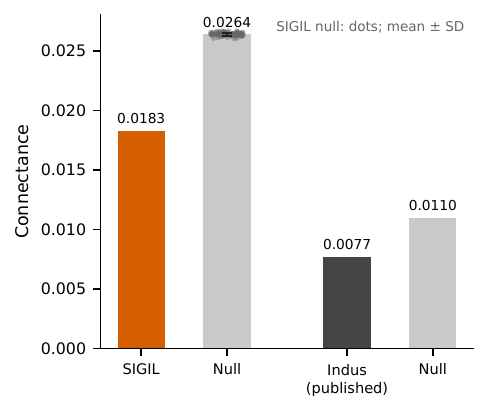}
\caption{Observed transition-graph connectance against the same stored null
draws. Only a small minority of the possible ordered sign pairs occurs in
either \sigil{} or the published Indus graph.}
\label{fig:connectance}
\end{figure}

\paragraph{Connectance (Figure~\ref{fig:connectance}).}
The 3,209 observed edges occupy 1.83\% of the $419^2$ possible ordered
pairs, against a null mean of 2.64\%, while the published Indus values are
0.77\% and 1.1\%. Inventory size, short texts, and constrained slots all
suppress the fraction of possible transitions that can ever be observed, in
either corpus.

\paragraph{Collocation and segmentation.}
Dunning's statistic exceeds the source threshold of $G^2=10.83$ for 381 of
the 3,209 observed bigram types, a count we leave descriptive because the
source specifies no multiple-testing correction. The strongest pairs have
direct constructional explanations: the deity Hare-Dawn followed by its
licensed epithet Radiant scores 2,253, an order of magnitude above the
next pairs at 310.7, 249.5, and 224.1, all of them deity-epithet
combinations, with an opener-deity pair close behind at 217.2. A trigram version of the same statistic
marks 498 of 5,687 observed trigram types as significant without
correction, led by opener, deity, and epithet sequences. High collocation
scores identify selective combination without identifying the linguistic
status of the units.

The staged segmentation reconstruction starts from 1,441 distinct strings
of length at least five, and the complete-text and frequent-combination
stages segment 79.81\% of them using 229 complete texts of length two, 465
of length three, and 572 of length four, together with 57 frequent edge
and 56 frequent middle combinations. The source does not give a
reproducible singleton-frequency cutoff, and every tested cutoff from 1 to
20 eventually renders all strings segmentable, with even the strictest
cutoff of 20 needing only 27 begin, 5 end, and 62 middle singleton units,
so we report the interval from 79.81\% to 100\% rather than selecting an
endpoint. Related reconstructions of recursive segmentation, which splits
each of the 2,753 distinct texts at its weakest adjacent edge, and of
shell-position entropy over ten positional bins are stored in the same
contract as declared implementations of their underspecified sources.

\paragraph{Unicity arithmetic.}
The cited calculation substitutes an effective inventory of 439 and an
assumed redundancy of $\rho=0.7$ into a unicity-distance formula, $d=N/\rho$,
adapted from \citet{shannon1949}, giving $d=627.1$, and the corpus of
13,717 tokens exceeds that distance almost 22 times over. The formula presupposes a
plaintext and a cipher model, neither of which exists for \sigil{}, and it
uses neither the observed text length distribution nor any recovered key. A
corpus can therefore exceed the resulting distance while remaining
non-linguistic, so the inequality cannot validate a reading on its own.

\paragraph{Repetition.}
Sproat's rule counts 438 instances in which a sign repeats somewhere within
its text, of which 283 are adjacent, for a ratio of 0.646, well above the 0.10
boundary of the source's non-linguistic class. Only 13\% of \sigil{} texts
contain any repeat at all. English words produce the opposite arrangement:
repeats occur in 57.3\% of sampled texts, yet only 376 of 111,424 repeated
instances are adjacent. The ratio therefore describes how repetition is
arranged rather than how often it occurs, and \sigil{} supplies known
non-linguistic reasons for adjacency in deity emphasis, double portions,
repeated epithet licenses, and tallies. A length-controlled reconstruction
of the same rule, applied to 500 texts truncated to at most six signs,
still yields a ratio of 0.815 with repeats in 9.8\% of texts, so text
length alone does not explain the classification.

\paragraph{Position-cell residuals.}
A separate descriptive analysis tests initial, medial, and final occurrence
counts
for 418 signs, excluding tokens that constitute a complete inscription by
themselves. The global statistic is $\chi^2=17{,}700.23$ on 834 degrees of
freedom, and 117 of 1,254 cells remain significant after Bonferroni
correction at a per-cell level of $3.99\times10^{-5}$, corresponding to a
two-sided $z$ cutoff of 4.11. The largest positive residual belongs to a
final terminal, and the other large cells contain openers and common role
signs. The test correctly rejects positional independence; its residuals
then recover the generator's slot inventory, not a grammatical category
system.

\paragraph{Classifier ingredients.}
The Lee coordinates combine unigram diversity, bigram diversity, singleton
counts, and token totals, so order enters only through the number of
distinct bigrams while much of the input survives shuffling. That is why
shuffled \sigil{} moves from the syllable to the word subclass but remains
writing, why the i.i.d. Zipf-Mandelbrot control is labeled ``writing:
words,'' and why the equiprobable control is labeled ``writing: letters.''
Only the rigid corpus falls below the writing threshold. Actual English
word sequences, evaluated over the retained top-400 corpus, land at
$C_r=83.9$ and $U_r=0.852$ and are therefore classified as letter writing,
so even a genuine language receives the wrong subclass. The rule separates
one narrow form of rigidity from several more varied processes; it does
not separate linguistic from non-linguistic generation for this comparison
class.

\paragraph{Feature panel and mean PMI.}
Sproat's comparative panel is reconstructed as a twelve-feature vector.
For \sigil{} it records $C_r=8.926$, $U_r=1.328$, a repetition ratio of
0.646, a maximum conditional entropy of 3.614 nats, and a mean pointwise
mutual information of $-5.87$ bits over the six sign types that carry the
top quarter of unigram mass. The original classification experiment
remains blocked, because its exact linguistic corpora, feature matrix,
split draws, and pruning path are not recoverable, so the panel is
reported as a described vector, not a classification.

\paragraph{Mutual information and directional diagnostics.}
The adjacent-sign mutual information is 2.114 bits from 10,717 unsmoothed
bigram tokens with boundaries excluded. Reconstructions of the recent
sequence diagnostics give a directional Kullback-Leibler divergence of
2.518 bits at order two and 1.238 bits at order three, forward against
reversed conditional entropies of 3.878 and 4.272 bits at order two, and
positional rigidity that is lowest in the second and third slots and rises
toward the ends of longer texts. The registry also records a source
conflict for the rigidity measure, whose published values exceed the range
of its stated formula.

\paragraph{Degree, strength, and significant edges.}
The degree and strength distributions show the same role structure from
another side. The largest in-degree is 129 and the largest out-degree 84,
while 127 signs, some 30\% of the inventory, never follow another sign and
only 13 never precede one. Under 10,000 within-text permutations, 502
edges exceed a $z$ score of 3 and 52 exceed 8, against published Indus
counts of 377 and 31, and the strongest edge, a deity-epithet pair,
reaches $z=40.4$. Selective combination is abundant; nothing in it
identifies the combining units as linguistic.

\paragraph{Frequency classes and inventory estimates.}
Fuls's frequency-class regression fits \sigil{} with a log-log slope of
$-1.06$ and $R^2=0.947$ over 84 frequency classes, and the source's
extrapolation rule would assign 71.3\% of the inventory to its syllabic or
single-consonant band; the registry keeps the result descriptive because
the typological calibration has no decision threshold. The three LNRE
population estimates likewise resist interpretation: the rejected finite
Zipf-Mandelbrot fit implies a population of only 699 signs, the rejected
GIGP fit implies more than four million, and the plain model supplies
none, so no estimate comes from an accepted model.

\paragraph{Fitted frequency parameters.}
Goodness of fit alone discriminates little in this family. The full
English word distribution fits the Zipf-Mandelbrot curve with $b=1.552$,
$c=440.9$, and $R^2=0.982$, the truncated top-400 English corpus with
$b=1.070$, $c=0.919$, and $R^2=0.997$, the independent control with
$b=1.686$, $c=6.06$, and $R^2=0.982$, and \sigil{} with $b=2.574$,
$c=28.46$, and $R^2=0.986$. All four are excellent fits with very
different parameters, produced by word use, truncation, independent
sampling, and reinforced urns respectively. The paper therefore treats
fitted parameters as description, not evidence.

\paragraph{Restoration benchmarks.}
A long-distance restoration benchmark reaches a top-one hit rate of 0.25,
a top-five rate of 0.61, and a top-ten rate of 0.67 with a mean reciprocal
rank of 0.389 over 100 evaluations, and a five-model panel over orders two
through seven peaks near a top-one rate of 0.27 for start-position filling.
Both are performance benchmarks in the registry rather than tests, because
neither source defines a linguisticity rule over its scores.

\paragraph{Estimator protocols.}
Several comparisons depend on declared caps that are easy to overlook.
The Brown character and word streams are capped at 300,000 tokens for the
block-entropy estimates, the Lee coordinates for English words are
computed over 269,711 contiguous runs holding 604,501 tokens of the
retained top-400 vocabulary, and the rigid and equiprobable controls
reduce to identical frequency spectra for the LNRE fits because both
realize 400 types over 200,000 tokens. Restoration fold means on the
90\%-mass criterion range from 83.6\% to 85.8\%, so the pooled 84.88\% is
stable across folds. Each of these choices is recorded beside its result
in the released contract.

\paragraph{Handedness controls.}
The terminal-asymmetry procedure behaves as designed on the controls,
which is what the comparison needs. Raw English characters receive a
significant
$\Delta G$ of $-0.175$, correctly read as left-to-right, while the
shuffled, i.i.d., and equiprobable controls show no significant
asymmetry. English words receive a small positive $\Delta G$ of $0.031$
that the source rule would read as right-to-left, a reminder that the
method measures terminal concentration, for which physical direction is
only one possible cause.

\paragraph{Null-draw bookkeeping.}
Every null comparison in this appendix runs on stored draws rather than
summary moments. The reciprocity and connectance nulls keep all 100
within-text permutations at a fixed seed, the edge-score analysis keeps
10,000, the $n$-gram maxima keep the ten draws their source used, and the
handedness tests keep 1,000 bootstrap and 1,000 shuffle draws apiece. A
rerun of the boundary-distribution protocol at an independent seed
reproduces the null means to three decimal places.

\paragraph{Analyses absent from this appendix.}
Ten registry entries could not be reconstructed, and their absence
is recorded, not repaired. The blocked entries lack, variously, a
recoverable source corpus, published thresholds and tie rules, released
code and hyperparameters, an archived feature matrix, or a site-level
protocol. Two entries set the pattern: the CART comparison lacks its
exact corpora, split draws, feature matrix, and pruning path, and the
recent BiLSTM perturbation study releases no code, splits, or
hyperparameters. Naming the gap keeps the census auditable and reopens
the entry when the artifact becomes available, without letting an
unreproducible method count either way.

\begin{figure*}[!htb]
\centering
\includegraphics[width=.80\textwidth]{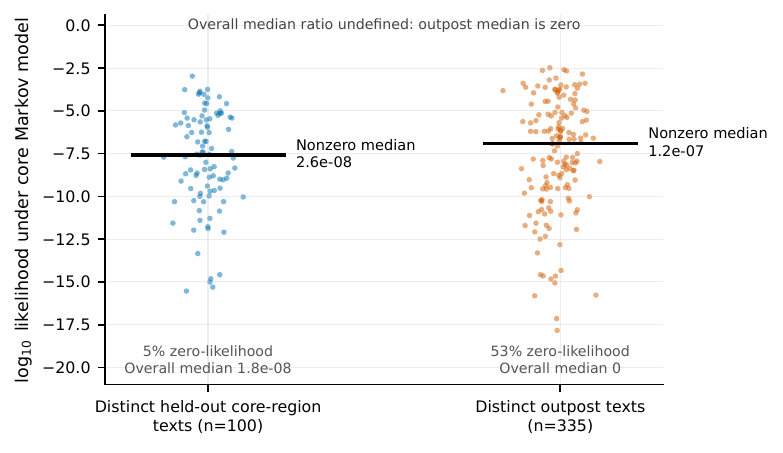}
\caption{Core-trained text likelihoods under the West Asia design, plotting
only texts with nonzero likelihood; the black bars are nonzero medians.
Overall medians, which include the zeros, are $1.85\times10^{-8}$ for
held-out core strings and exactly zero for the outpost, with zero shares of
5.0\% and 53.1\%. Published Indus medians are $1.12\times10^{-7}$ and
$6.4\times10^{-13}$.}
\label{fig:west-asia}
\end{figure*}

\paragraph{Outpost likelihood (Figure~\ref{fig:west-asia}).}
Among nonzero scores the held-out median is $2.57\times10^{-8}$ and the
outpost median is $1.18\times10^{-7}$, so restricting attention to positive
values would reverse the impression left by the overall medians. The
complete distribution shows that the separation is driven almost entirely by
unseen outpost events, as the declared regional mechanism predicts, and a
reader given only a median ratio would never see that structure.

\begin{figure}[!htb]
\centering
\includegraphics[width=\columnwidth]{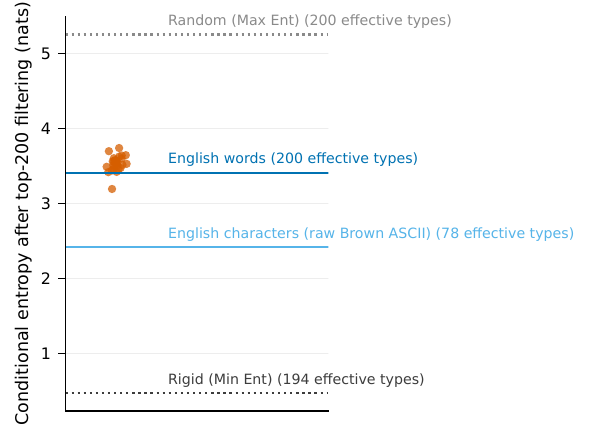}
\caption{Conditional entropy after top-200 filtering for the common-seed
27-cell sensitivity grid, which varies urn concentration, optional-slot
probability, and epithet-license length. Reference lines show the fixed
linguistic, random, and rigid comparisons at their own effective
inventories.}
\label{fig:sensitivity-ce}
\end{figure}

\paragraph{Entropy sensitivity (Figure~\ref{fig:sensitivity-ce}).}
Across the grid, conditional entropy at $k=200$ ranges from 3.195 to 3.741
nats, and the central cell reproduces the canonical corpus text for text.
The cells share a starting seed while parameter-dependent control flow
makes their random streams diverge. We therefore present the grid as a
sensitivity display, and paired uncertainty claims would be inappropriate.

\begin{figure}[!htb]
\centering
\includegraphics[width=.88\columnwidth]{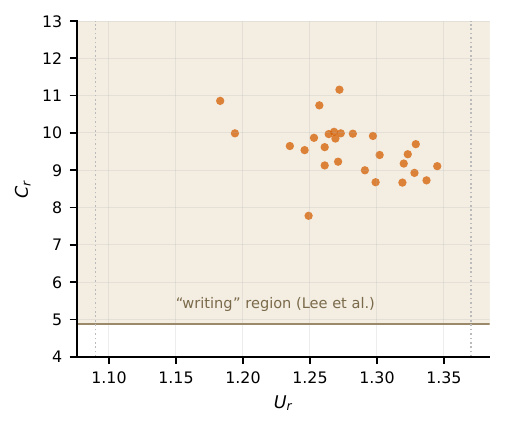}
\caption{All 27 sensitivity configurations exceed the Lee writing threshold
and fall inside its syllabic subclass.}
\label{fig:sensitivity-lee}
\end{figure}

\paragraph{Classifier sensitivity (Figure~\ref{fig:sensitivity-lee}).}
Every configuration is labeled ``writing: syllables,'' with $C_r$ between
7.78 and 11.16 and $U_r$ between 1.183 and 1.345. This is robustness within
one constructed architecture, and we present it as exactly that rather than
as validation of the classifier on an independent sample of emblem systems.

\section{Sequential Reconstruction Results}\label{app:decoder}

\paragraph{Engine and budgets.}
The shared engine is deterministic end to end, and every budget is
declared in the released contract. Candidate propagation uses a beam of
width four over ten steps with at most six candidate values per sign,
equality-pattern evidence comes from up to 192 short or repeated
inscriptions whose signs number at most ten, candidate sets are capped at
256 entries, and one refinement pass revisits up to 72 signs. Up to five
segmentations are stored per text, and segmentation counts are enumerated
exactly under a cap of $10^9$ that no text ever reaches. Four canonical
deterministic starts, twenty random keys, six bootstrap resamples, and the
grouped split are all seeded, so every number in this appendix can be
regenerated bit for bit.

\paragraph{Lexicon builds.}
The English lexicon is the pinned NLTK word list, retaining 196,943 forms
of at most twelve letters with one-letter entries restricted. The Sanskrit
lexicon is a pinned digitization of the Monier-Williams dictionary
\citep{monier1899}: 194,015 eligible headwords yield 164,943 exact SLP1
forms at the literal level, and the script protocol expands them to
2,664,046 finite forms through thirteen mechanical a-stem rules and two
final sandhi alternants, with ten further suffix rules available only to
the adaptive overlay. The Tamil lexicon holds 51,009 forms from the
wordfreq frequency list \citep{speer2022wordfreq}, with Tamil-script
entries transliterated and folded through Aksharamukha
\citep{aksharamukha}. Every input is verified by content digest before
use, and none supplies glosses to the search.

\paragraph{Protocol nesting and hygiene.}
The three Sanskrit protocol levels are verified strict subsets: every
literal reading remains available at the script level, and every script
reading at the adaptive level. All languages share the same equality
regexes, intersection, beam propagation, complete segmentation, split
rules, and budgets, and the Sanskrit-specific writing freedoms are never
transferred to English or Tamil. The adaptive level may add up to 64
dictionary entries but in fact adds none; its eleven logged interventions
are all downstream key revisions, and the held-out adaptive runs record no
training-only additions of any kind. Duplicate isolation across the
grouped split is checked explicitly, and \sigil{} semantics are supplied
to no stage of the search.

\paragraph{Source constraints.}
The classification of this experiment as a source-constrained
reconstruction rests on five recorded limitations of its source. The
released Perl code processes manually written regular expressions rather
than generating them, its test corpus covers 32 sign headings rather than
a complete signary, its search order is underspecified beyond
illustrative short inscriptions and frequency guidance, its beam width,
dead-end revision, tie rules, segmentation enumeration, and held-out
evaluation are absent, and its dictionary download is unpinned with
manually added auxiliary forms. Our implementation answers each gap with
a declared choice, uses a pinned dictionary, and separates every adaptive
addition into its own logged level, so nothing that the source leaves
open is silently resolved in a favorable direction.

\paragraph{Search-space bookkeeping.}
The whole-word constraint stage generates 12,623 candidate values for
English, 24,710 for literal Sanskrit, 28,264 at the script and adaptive
levels, and 22,541 for Tamil. Each complete key table has 450 rows, one
per registered identity, and the 31 identities unattested in the core
corpus carry an explicit unresolved status, so no key ever reports more
values than the corpus can support. Every table, split, and control is
identified by content digest in the contract, and the published seeds
cover the four deterministic starts, the twenty random keys, the six
bootstrap resamples, the grouped split, and both perturbed-text controls.

\paragraph{Determinism and outputs.}
Each protocol stores its complete key table, the first forty provisional
assignments in search order, a logged derivation trace, per-text
segmentation-ambiguity arrays, the identifiers of its readable texts, and
a SHA-256 digest of the key itself. The controls receive the same
treatment, with the twenty random keys and six bootstrap keys stored
beside their own tables, so a control run can be audited exactly like
the canonical one. Eight further digests pin the
inputs, covering the decoder source, the lexicon builder, the raw
dictionary and word-list downloads, the derived Tamil ranks, the
portable corpus, the upstream provenance, and the corpora artifact, and
the release refuses mismatched inputs rather than warning about them.
The decipherment experiment therefore cannot silently drift away from
the corpus it claims to read, and any independent rerun can confirm
agreement down to the byte rather than to a rounded percentage.

\begin{figure*}[!htb]
\centering
\includegraphics[width=.82\textwidth]{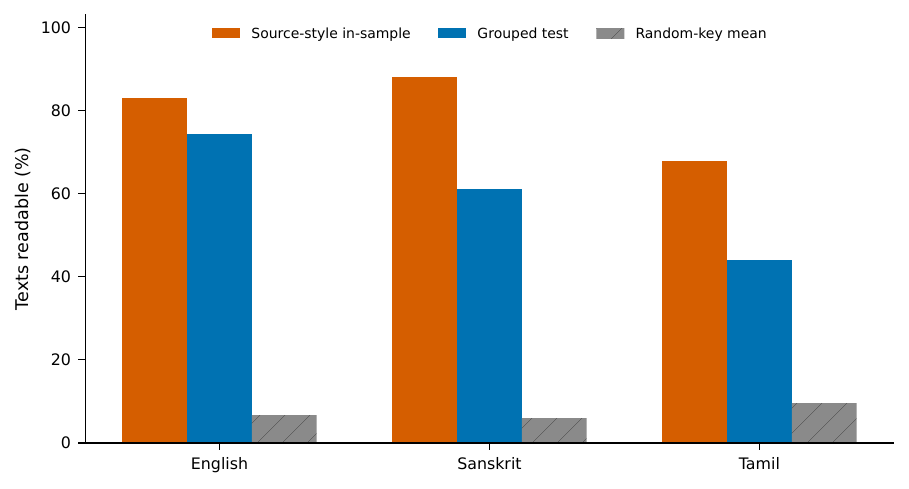}
\caption{Literal-protocol parse coverage for English, Sanskrit, and Tamil
under one shared sequential decoder. Source-style values are in-sample over
all 3,000 texts; grouped-test values are computed on the 454-text held-out
partition after duplicate isolation; hatched bars show means over 20 random
keys.}
\label{fig:multilingual}
\end{figure*}

\paragraph{Multilingual coverage (Figure~\ref{fig:multilingual}).}
The three languages matter jointly, because they succeed on the same input.
One corpus without phonological values admits high optimized coverage under
three unrelated lexicons, which is only possible if the coverage statistic
measures search freedom and lexicon geometry alongside anything about the
corpus. Grouped evaluation lowers every language, most sharply Sanskrit and
Tamil, but a lower feasibility number is still a feasibility number: nothing
in the decline converts coverage into identification. The English list
contains 196,943 retained forms, the Sanskrit literal lexicon 164,943, and
the Tamil list 51,009.

\paragraph{Where the freedom enters.}
The literal keys map the 419 attested sign identities into alphabets of 26
English letters, 49 Sanskrit units, or 19 transliterated Tamil letters, so
homophony is unavoidable everywhere and extensive in English and Tamil. Up
to 192 short or repeated inscriptions contribute equality-pattern
constraints. A sign value counts as resolved only when every applicable
candidate set leaves a single non-conflicting intersection, a standard met
by seven English values, one literal Sanskrit value, and no Tamil value;
every other
entry in the released key tables is a provisional search hypothesis and is
labeled as one.

The search objective prefers, in order, more completely readable texts, more
tokens inside those texts, more mechanically analyzed forms, more
independently attested forms, fewer relaxations, fewer dictionary additions,
and fewer unresolved signs, with a deterministic tie break at the end. These
priorities make every tie reproducible, and they also make the nature of the
score explicit: the target is dictionary segmentation, and no term anywhere
in the objective rewards recovery of the known \sigil{} meaning.

\paragraph{What the controls say.}
The random-key means in Figure~\ref{fig:multilingual} isolate the
contribution of search. Two further controls hold the optimized literal key
fixed and replace the texts instead. Within-text shuffles remain readable at
72.30\% in English, 86.03\% in Sanskrit, and 47.27\% in Tamil, and
length-preserving i.i.d. unigram samples reach 67.90\%, 86.03\%, and
39.33\%. Because the key is never refitted, these values show how much of
the Boolean segmentation rate is already explained by short texts, unigram
concentration, and the lexicons themselves. They are diagnostics of the
score, not alternative decipherments.

The grouped split contains 2,115 training texts, 431 validation texts, and
454 test texts, holding 9,692, 1,930, and 2,095 tokens respectively, and
exact duplicate strings never cross those partitions. Shared signs and
shorter motifs remain available across the split, as they would in any
corpus, so the grouped test probes transfer to unseen complete
inscriptions without pretending that their components are independent.

\paragraph{Selected readings and traces.}
The contract stores four reading panels per language: up to 400
deterministically chosen medium-length texts of 2,060 eligible, the fifty
longest texts, the fifty shortest, and up to one hundred of the 256
ligature-bearing texts. Each row records the rendered string, its
dictionary segmentation, alternative readings, and the true \sigil{}
semantic record, which is attached only after every key and parse is
fixed. Fluent translation is marked unavailable throughout, because the
pinned dictionary entries do not license a compositional rendering. The
stored English derivation trace shows how the search actually proceeds:
its twelfth step assigns one deity sign the value \emph{i} with an
information gain of 14.1 bits, after which 1,131 texts are readable, and
two later logged steps raise that count to 1,515.

\paragraph{Perturbed-text denominators.}
One denominator subtlety deserves a note. The fixed-key shuffled and
i.i.d. controls for the Sanskrit script and adaptive protocols decompose
the seventeen genuine ligatures into their components, so their token
denominator is 13,981 rather than 13,717, while all text-level shares
remain out of 3,000. The stored contract keeps both denominators explicit
so that no percentage silently mixes the two.

\paragraph{English and Tamil diagnostics.}
The diagnostics that Appendix~\ref{app:decoder-diagnostics} plots for
Sanskrit exist for the other languages as well, computed from the stored
per-text arrays. English readable texts admit a median of two dictionary
segmentations with a maximum of 89, and 39.0\% have a unique parse; Tamil
readable texts have a median of one, a maximum of 20, and 68.1\% unique
parses. The six bootstrap keys span full-corpus coverage from 64.7\% to
78.8\% in English and from 50.3\% to 60.9\% in Tamil, and the four
deterministic starts return readable-text counts from 2,435 to 2,490 in
English and from 1,858 to 2,032 in Tamil with mean pairwise key agreement
of 13.4\% and 11.5\%. Frequency-stratified agreement repeats the Sanskrit
pattern: signs observed at most twice agree at 2.8\% in English and 5.3\%
in Tamil, while signs observed at least twenty times agree at 39.2\% and
34.2\%.

\paragraph{A reading in miniature.}
The shortest texts expose the mechanics plainly. One single-sign
inscription, whose emblem is the deity Ibex-Flood, is read in Sanskrit as
the bare seventh vowel, whose Monier-Williams entry is simply the name of
that vowel, and in English as the one-letter word \emph{i}; both count as
fully readable texts under complete segmentation. The panels mark every
such reading's fluent translation as unavailable, because the dictionary
entries license no compositional rendering, and they attach the true
administrative meaning only after the key is fixed. A reader can browse
four hundred of these rows per language and watch high coverage coexist
with arbitrary values sign by sign.

\paragraph{Attested-form accounting.}
The in-sample counters make visible where each protocol's coverage comes
from. Readable English texts contain 8,205 independently attested
dictionary forms and use no phonological relaxations at all, as the
literal rules require. Literal Sanskrit contains 10,794 attested forms,
again with no relaxations, and readable Tamil texts hold 4,260 attested
forms among 9,422 decoded units under the same literal discipline. The
script level contains 11,368 attested
forms alongside 1,781 mechanically generated morphological analyses and
3,662 applications of the declared phonological freedoms, and the
adaptive level raises those to 2,210 analyses and 4,626 relaxation uses,
while dictionary additions remain at zero everywhere. Unit totals give
the same picture from above: literal Sanskrit decodes its readable texts
into 11,946 units while the script and adaptive levels reach 13,551 and
13,588, so the relaxed protocols admit roughly 1,600 additional units'
worth of texts, and the growth is carried by mechanical analyses and
phonological freedoms rather than by newly attested vocabulary. On grouped
validation partitions, literal, script, and adaptive coverage reach
58.2\%, 94.7\%, and 96.1\%, tracking the corresponding test values
closely. The accounting shows that the relaxed protocols add ways to
succeed without adding evidence.

Two consistency checks round out the accounting. In-sample token-level
readable shares track the text-level shares within about a point, while
the held-out literal profiles open gaps of up to five and a half points,
so the contract reports both denominators wherever they differ. And
validation and test coverage sit within three percentage points of each
other at every Sanskrit level, evidence that the
grouped protocol is not being tuned against its own test partition, while
the much larger gap to the in-sample values isolates what full-corpus
optimization contributes. Both checks can be recomputed from the stored
per-text arrays alone, without rerunning any search or refitting any
key, and the arrays are indexed by text identifier, so any single row
can be spot checked directly.

\begin{figure*}[!htb]
\centering
\includegraphics[width=.82\textwidth]{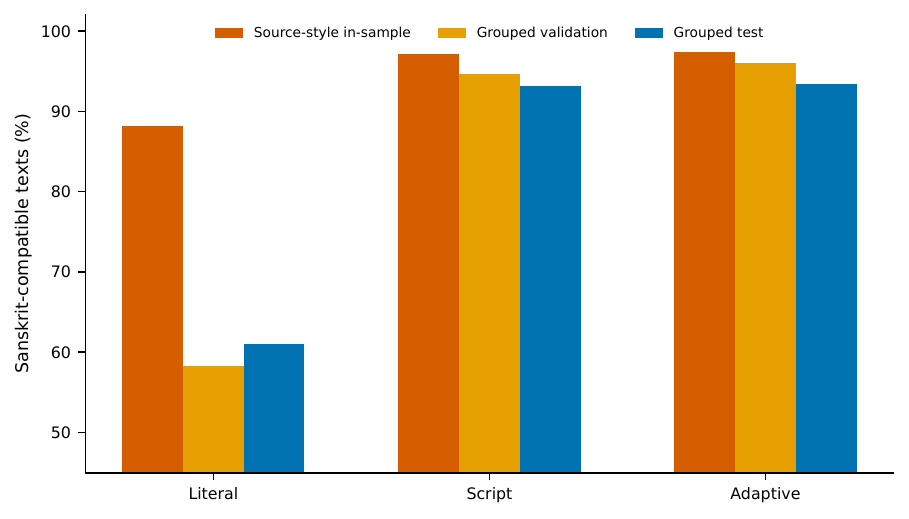}
\caption{Sanskrit coverage at the literal, script, and adaptive protocol
levels, evaluated source-style in-sample and with a training-only key on the
grouped validation and test partitions. The vertical axis begins at 45\%.}
\label{fig:sanskrit-protocols}
\end{figure*}

\paragraph{Protocol latitude (Figure~\ref{fig:sanskrit-protocols}).}
Literal held-out performance falls far below its in-sample value, and the
script level closes most of that gap. The reason is capacity, not
correctness: the script lexicon contains 2,664,046 mechanically generated
forms, and its declared writing freedoms multiply the realizations of every
key. The adaptive level then adds only a small increment, contributing 11
logged downstream key revisions and no dictionary additions. Keeping the
levels separate locates where the coverage comes from instead of merging
every freedom into a single headline number.

\begin{table}[H]
\centering
\small
\begin{tabular}{@{}lL{.31\columnwidth}L{.24\columnwidth}@{}}
\toprule
Profile & Coverage: in / test / random & Evidence: singleton / agreement \\
\midrule
English literal & 83.00 / 74.23 / 6.50 & 7 / 11.33 \\
Sanskrit literal & 88.13 / 61.01 / 5.77 & 1 / n.a. \\
Sanskrit script & 97.10 / 93.17 / 27.87 & 0 / 9.30 \\
Sanskrit adaptive & 97.33 / 93.39 / 27.87 & 0 / n.a. \\
Tamil literal & 67.73 / 44.05 / 9.48 & 0 / 12.09 \\
\bottomrule
\end{tabular}
\caption{Sequential reconstruction outcomes. Random-key values are means
over 20 draws. Bootstrap agreement is mean pairwise exact key agreement over
six training resamples, reported for each language's primary stability
protocol. All coverage and agreement entries are percentages.}
\label{tab:decoder}
\end{table}

Table~\ref{tab:decoder} separates three questions that a single coverage
percentage would conflate. In-sample feasibility asks what an optimized
key can fit, grouped-test feasibility asks what transfers to unseen
inscription types, and the evidence columns ask whether particular sign
values are identified at all. The three questions receive progressively
weaker answers, and the weakest of them is the one a decipherment claim
actually needs. The random-key column calibrates each profile's floor,
which ranges from 5.8\% to 27.9\% as the protocols relax, and the
agreement column shows that even the best-covering protocol identifies
almost nothing at the level of individual signs.

\section{The Literature Registry}\label{app:registry}

The registry search was defined before any scoring ran. Four declared query
families across Crossref, Google Scholar, the ACL Anthology, and publisher
full-text search, seven backward-citation seeds, and a forward-citation
screen running to the close of the census delimit its contents. Its scope is
peer-reviewed, language-independent corpus and sequence analyses applied to
an Indus corpus, with declared exclusions for methods that depend on
archaeological metadata, glyph shapes, or an assumed reading, for proposed
decipherments, and for work not peer reviewed by the census close. The method is
the unit of entry, so a paper reporting several formulas or decision rules
contributes several entries, and the registry stores one status per
method with its reason, retaining source-validation failures instead of
discarding them.

\begin{table}[H]
\centering
\small
\begin{tabular}{@{}llL{.52\columnwidth}@{}}
\toprule
Status & Count & Basis \\
\midrule
Exact scored & 6 & Reproduced Indus outcome and source decision rule \\
Reconstruction & 18 & Consequential source details declared here \\
Descriptive & 14 & Implemented quantity without a categorical rule \\
Blocked & 10 & Required data, mapping, threshold, or target absent \\
Out of scope & 6 & Does not test statistical language specificity \\
\bottomrule
\end{tabular}
\caption{The complete status partition of the 54-method registry at the
census close. Only the first row contributes to the exact comparison. The
machine-readable contract enumerates every method with its citation,
protocol, and status rationale, together with source hashes where a public
resource exists.}
\label{tab:registry}
\end{table}

Table~\ref{tab:registry} guards against two opposite forms of selective
reporting at once. Weakly documented methods cannot silently enlarge the
denominator, and methods that could not be scored cannot disappear from
view. The qualified claim licensed by the registry is therefore ``all exact
source-validated outcomes in the registry,'' never ``all published
analyses.''

The entries span the families a reader of the debate would expect. The
entropy and Markov analyses of Rao and colleagues
\citep{rao2009science,rao2009pnas} appear beside the $n$-gram, positional,
clustering, and restoration methods of Yadav and colleagues
\citep{yadav2010plos,yadav2008segmentation,yadav2008pattern,yadav2017clusters},
the network and core analyses of Sinha and colleagues
\citep{sinha2011network,ashraf2012core}, the handedness procedure
\citep{ashraf2018handedness}, and the LNRE and positional tests of
\citet{oakes2019}. Sproat's comparative feature panel and repetition ratio
\citep{sproat2014language}, Fuls's classification measures \citep{fuls2015},
segmentation methods reaching back to \citet{siromoney1988}, early
syntactic association analyses \citep{koskenniemi1981}, geolocation
prediction \citep{recchia2016},
restoration and West Asia models
\citep{venkatesh2023,venkatesh2023westasia}, dependency clustering
\citep{ramani2024}, sign-function analyses \citep{khanna2025}, recent
sequence models \citep{tiwari2026}, a contemporaneous synthetic-baseline
preprint \citep{nair2026}, and Farmer, Sproat, and Witzel's corpus-level
observations \citep{farmer2004} complete the census. Blocked entries
typically lack public data or a defensible common target, and the
out-of-scope entries include methods defined only on other symbol corpora
and methods that presuppose a sign reading. The sequential Sanskrit proposal
is not a registry entry, because the declared exclusions remove proposed
decipherments and work not peer reviewed by the census close; it is evaluated
separately through the source-constrained reconstruction of
Section~\ref{sec:decoder-method} and Appendix~\ref{app:decoder}.

\begin{figure}[!htb]
\centering
\includegraphics[width=.78\columnwidth]{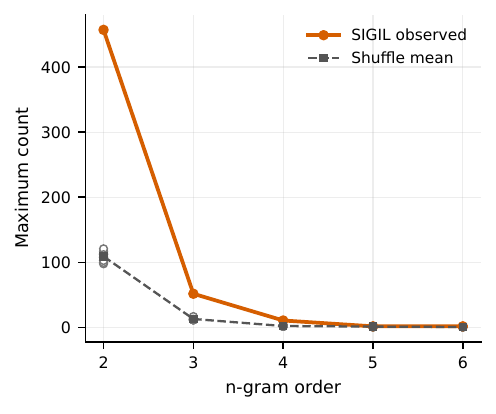}
\caption{Maximum $n$-gram counts against ten stored length-preserving
within-text shuffles per order. The method is a declared reconstruction and
does not enter the exact score.}
\label{fig:ngram-null}
\end{figure}

\paragraph{$n$-gram maxima (Figure~\ref{fig:ngram-null}).}
The most frequent observed bigram occurs 457 times against a shuffle mean of
109.3, and the trigram maximum is 52 against 13.3. A difference persists at
order four, the order-five maximum is unexceptional, and the order-six
maximum sits just above its null. The ten stored draws match the source
count exactly, and ten draws are also too few for precise tail inference,
which is one more reason this entry remains a reconstruction.

\begin{figure}[!htb]
\centering
\includegraphics[width=.78\columnwidth]{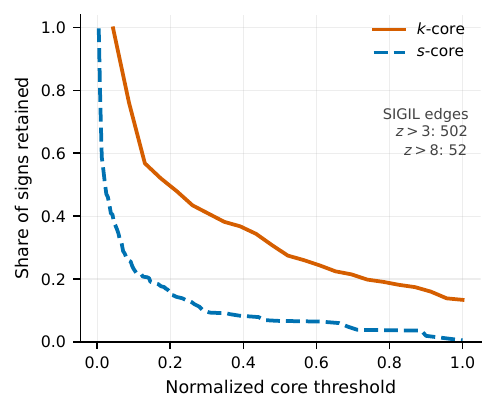}
\caption{Undirected $k$-core and weighted $s$-core retention as the
normalized threshold rises, with source-threshold edge counts from the
reconstructed network null in the annotation.}
\label{fig:network-cores}
\end{figure}

\paragraph{Network cores (Figure~\ref{fig:network-cores}).}
Recursive pruning yields nested core sets, and the deepest nonempty
undirected $k$-core has threshold 23 while retaining 56 signs. Directed in-
and out-core variants differ from each other because block order is
directional. Structures of this kind are expected wherever frequent role
signs link into specialized tails, and \sigil{} contains exactly such role
signs by construction.

\begin{figure}[!htb]
\centering
\includegraphics[width=.78\columnwidth]{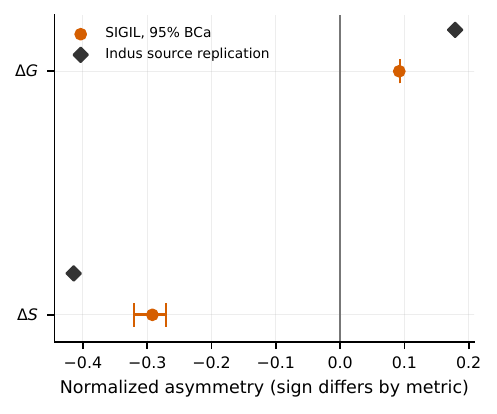}
\caption{Terminal handedness. Circles and whiskers are \sigil{} point
estimates with 95\% BCa intervals; diamonds are the reproduced Indus
estimates. Both metrics consume the normalized spatial order.}
\label{fig:handedness}
\end{figure}

\paragraph{Handedness (Figure~\ref{fig:handedness}).}
For \sigil{}, $\Delta G=0.091995$ and $\Delta S=-0.291969$, and each
two-sided randomization test gives $p=1/1001$. The stored 95\% BCa
intervals for both asymmetries exclude zero, as the source's rule
requires. The source rule infers a
right-to-left convention, which is correct for the normalized corpus, so
the procedure succeeds at the task it was built for. What the outcome
demonstrates is that handedness and linguistic status are separate targets,
and that only the first is being measured.

\begin{figure}[!htb]
\centering
\includegraphics[width=.78\columnwidth]{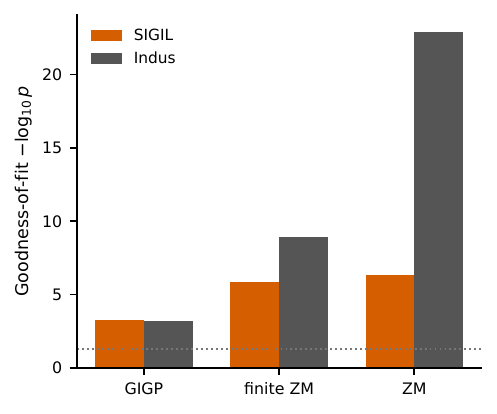}
\caption{LNRE goodness-of-fit tests for \sigil{} and the reproduced Indus
data, shown as $-\log_{10}p$ with the dotted line at $p=0.05$; taller bars
mean poorer fit.}
\label{fig:lnre}
\end{figure}

\paragraph{LNRE models (Figure~\ref{fig:lnre}).}
The SciPy port reproduces the source's Indus goodness-of-fit rejections from
the published frequency spectrum \citep{oakes2019}, and all three models are
rejected on \sigil{} as well. Poor fit to a single global lexical frequency
law is unsurprising for an emblem inventory whose categories draw from
separate urns and whose issue marks are bounded by quota, which is precisely
the structure \sigil{} declares.

\section{Sequential Diagnostics}\label{app:decoder-diagnostics}

\begin{figure}[!htb]
\centering
\includegraphics[width=.78\columnwidth]{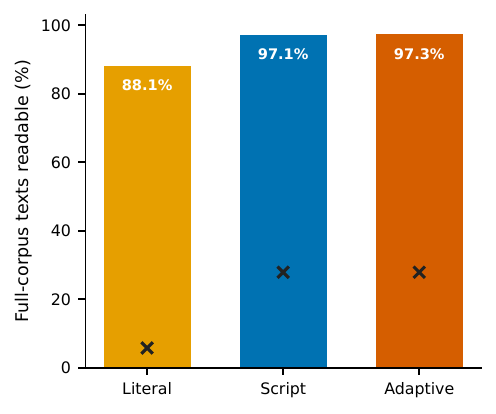}
\caption{Full-corpus Sanskrit feasibility at the three protocol levels, with
crosses giving random-key means under the same dictionaries and writing
rules.}
\label{fig:yd-readability}
\end{figure}

\paragraph{Full-corpus readability (Figure~\ref{fig:yd-readability}).}
Literal, script, and adaptive coverage reach 88.13\%, 97.10\%, and 97.33\%,
while the corresponding random-key means are 5.77\%, 27.87\%, and 27.87\%.
Optimization clearly raises coverage, but the script-level random baseline
nearly quintuples the literal one, which shows how much feasibility flows
from the expanded form set before any search begins.

The gaps between the optimized and random values tell the sharper story.
At the literal level, search lifts coverage from 5.77\% to 88.13\%, a
factor of about fifteen, while at the script and adaptive levels the
lift runs from 27.87\% to 97.10\% and 97.33\%, a factor of about three
and a half. The freer the representation, the less work optimization has
left to do, because the expanded form set already parses over a quarter
of the corpus under arbitrary sign values. The adaptive freedoms add
nothing under random keys, so the script and adaptive baselines coincide
at 27.87\%, and the two optimized values differ by less than a quarter
of a point. The panel therefore splits the headline coverage into what
the lexicon grants for free, what the declared freedoms grant, and the
remainder that search must earn.

\begin{figure}[!htb]
\centering
\includegraphics[width=.78\columnwidth]{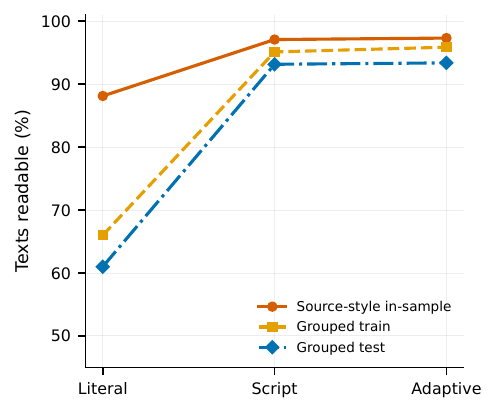}
\caption{Sanskrit source-style, grouped-training, and grouped-test coverage
at each protocol level. The held-out key and any adaptive overlay are
learned from training groups only, and the vertical axis begins at 45\%.}
\label{fig:yd-heldout}
\end{figure}

\paragraph{Held-out behavior (Figure~\ref{fig:yd-heldout}).}
Literal coverage reaches 66.00\% on training groups and 61.01\% on test
groups, against 88.13\% under full-corpus optimization, while script-level
values are 95.13\% and 93.17\% and adaptive values are 95.89\% and 93.39\%.
Duplicate isolation prevents exact inscriptions from leaking across the
split, although repeated substructures and frequent signs remain shared, as
they would in any real corpus.

\begin{figure}[!htb]
\centering
\includegraphics[width=.78\columnwidth]{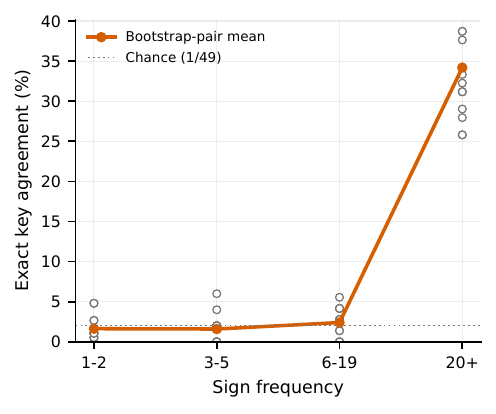}
\caption{Exact pairwise agreement among six Sanskrit script keys learned
from grouped bootstrap samples, stratified by sign frequency. Open points
are the fifteen key pairs; the line gives bin means; the dotted line marks
the 1-in-49 chance rate.}
\label{fig:yd-stability}
\end{figure}

\paragraph{Key stability (Figure~\ref{fig:yd-stability}).}
Overall agreement is 9.30\%. The 187 rarest signs, each observed once or
twice, agree at 1.64\%, below the nominal chance rate of one in 49, and the
93 signs observed at least 20 times agree at 34.19\%. All six keys
nevertheless parse more than 93.5\% of the full corpus. Coverage is stable
while the purported readings underneath it are not. That pattern indicates
a weakly identified value system, not a recovered one.

\begin{figure}[!htb]
\centering
\includegraphics[width=.78\columnwidth]{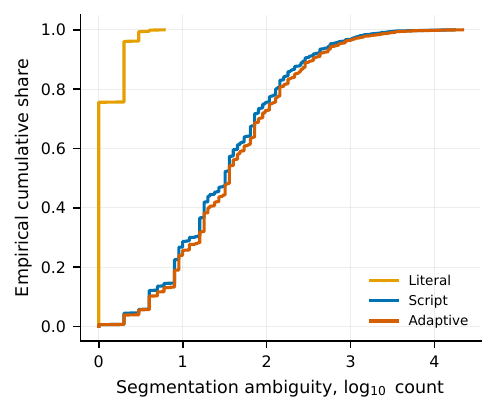}
\caption{Empirical cumulative distributions of Sanskrit segmentation counts
over each protocol's readable texts, with counts floored at one before the
logarithm.}
\label{fig:yd-ambiguity}
\end{figure}

\paragraph{Ambiguity (Figure~\ref{fig:yd-ambiguity}).}
Readable strings often admit several dictionary segmentations at once. The
literal protocol stays modest, with a median of one segmentation per
readable text and a maximum of six, while the script and adaptive protocols
reach medians of 32 and 36 and maxima of 17,280 and 21,600; the enumeration
cap of $10^9$ is never approached. Each protocol has a unique best run
within the three-text tolerance, yet the four deterministic starts return
four different keys, with mean pairwise agreement of 12.33\% for literal,
12.23\% for script, and 13.06\% for adaptive. A deterministic winner does
not make the value system well identified.

\begin{figure}[!htb]
\centering
\includegraphics[width=.82\columnwidth]{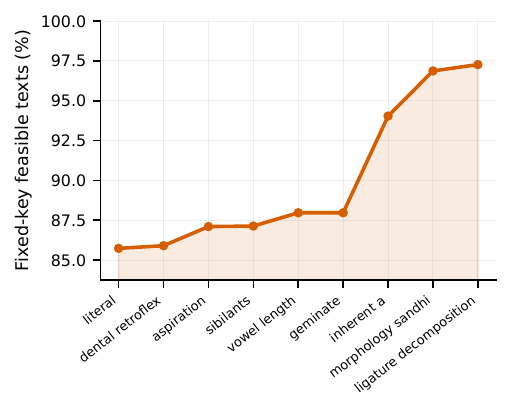}
\caption{Cumulative fixed-key Sanskrit feasibility as script freedoms are
added in the declared order, starting from the fixed-key literal baseline of
85.73\%. The final stage admits optional decomposition of the 17 genuine
\sigil{} ligatures.}
\label{fig:yd-relaxations}
\end{figure}

\paragraph{Relaxation ablation (Figure~\ref{fig:yd-relaxations}).}
Dental-retroflex, aspiration, sibilant, vowel-length, and geminate handling
together move fixed-key coverage from 85.73\% to 87.97\%. Optional inherent
\emph{a} contributes the single largest step, reaching 94.03\%, mechanical
morphology and sandhi reach 96.87\%, and optional ligature decomposition
ends at 97.27\%. Because the key never changes across the ladder, the
ablation isolates representational latitude from search, and it attributes
most of the script-level gain to the representation.

One definitional detail belongs on the record. The ladder's final stage
counts a text as feasible if either its written or its decomposed form
parses, reaching 2,918 texts, while the script protocol's headline counts
decomposed views alone and reaches 2,913; the contract stores both
numbers under separate keys so the two definitions cannot be conflated.
The same care applies to the fixed-key literal baseline of 85.73\%, which
sits below the searched literal result of 88.13\% because no key
optimization is permitted inside the ablation.

\paragraph{Semantic recovery.}
The contract stores the semantic-recovery outcome as a blocked method,
with no score and no synthetic permutation null. Monier-Williams
definitions are heterogeneous prose, while \sigil{} meanings are
structured records of deities, offices, goods, tallies, and validation,
and no predeclared ontology links the two domains. Deciding after the
fact which dictionary senses correspond to which administrative roles
would manufacture the very target being tested, so the contract marks
the method blocked and states the reason rather than reporting a
favorable or unfavorable number.

\section{The Complete Signary}\label{app:signary}

The signary is an invention; the functions it bundles are not. Each
category does work that the archaeological record attests for Bronze Age
seals and emblems, so a polity of that age could have operated a system
of this shape without encoding a word of speech. The registry holds 450
immutable identities: 6 openers, 242 deities, 60 epithets, 48 guilds, 4
ranks, 40 commodities, 8 tallies, 20 issue marks, 5 terminals, and 17
ligatures, the deity count including the two outpost cults.

The devotional face follows Near Eastern seal practice, where a legend
names a patron deity, may attach an epithet, and identifies the owner by
service or office \citep{collon1987}. The six openers name auspicious
occasions in the manner of dedicatory formulae, deities pair an animal
or natural emblem with a celestial or elemental term, and the 60
epithets are single honorifics whose weighted licenses attach to 241 of
the 242 deities, so recurring deity and epithet pairs follow from the
licenses. Readings of Indus inscriptions as formulaic, non-phonological
conveyance of meaning describe a system of exactly this kind
\citep{mukhopadhyay2019}.

The economic face is archaic bookkeeping. Proto-cuneiform accounts pair
commodity signs with discrete numeral notations \citep{englund2011}, and
Mesopotamian sealing bound such transactions to officeholders inside
institutional hierarchies \citep{zettler1987}. Here, 40 commodities take
one of 8 tallies counting one through eight, and 48 guilds take one of
4 ranks; a brickmaker of rank one entitled to one measure of a numbered
good is a routine text, and the kind of record a sealing archive exists
to keep.

The control face has narrower analogues. Late Bronze Age Anatolian pot
marks form small bounded repertoires tied to administrative control
\citep{glatz2012}, and the 20 issue marks work the same way, four
regions crossed with five rounds under bounded service quotas. The five
terminals condense the validating act of sealing itself
\citep{zettler1987}. Harappan seals differ by region within one
interacting world \citep{ameri2013}, and the four core regions likewise
carry their own patrons and slot conventions, while the outpost follows
the Gulf pattern of transplanted sealing practice mixing inherited and
local elements \citep{laursen2010}, shared deities beside two local
cults.

None of this grounding enters the statistics. Every identity carries an
integer ID, a category, and a display gloss, sealed after generation,
and the renderer derives each glyph from category and ID alone, so no
gloss or cosmetic stroke can change any analysis. Of the registered
signs, 419 occur in the core corpus and 166 at the outpost; 268 types
are core-only and 15 outpost-only, twelve of them deities including the
cults, and sixteen epithets are never realized at all, differences the
regional likelihood experiment measures rather than collapses. The plate
that follows is read cell by cell, glyph above gloss under a category
heading with its count, and 434 of the 450 identities are attested
across the two corpora. Glyph construction follows a fixed grammar, one
silhouette family per category with strokes varied by identity number,
and two complete rebuilds of the pipeline produce byte-identical plates.
The plate lists the categories in slot order, openers first and
ligatures last, so its headings retrace the template of a text.

\begin{figure*}[p]
\centering
\includegraphics[height=.92\textheight]{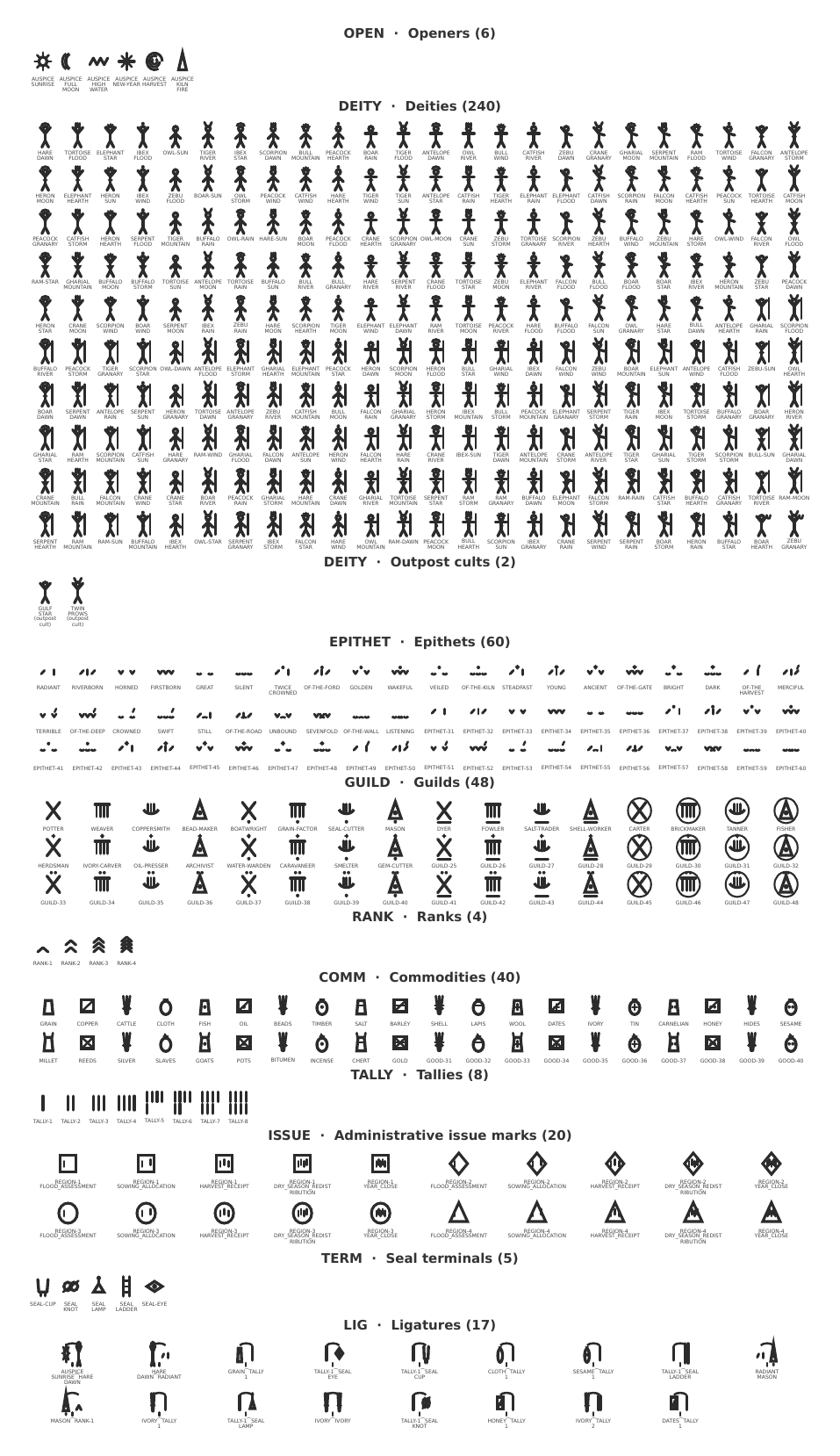}
\caption{All 450 registered \sigil{} identities. Category headings and
glosses state the constructed function of every emblem.}
\label{fig:signary-all}
\end{figure*}

\paragraph{Overview (Figure~\ref{fig:signary-all}).}
The full chart doubles as an inventory check and as a picture of the
unequal category sizes. Deities dominate the type count while a small set
of terminals and tallies contributes a large share of recurring tokens,
and that type-token imbalance is a direct source of both the long
frequency tail and the positional concentration measured earlier. The
inventory also sits where an excavated one would: 419 attested types
among 450 registered identities, beside the 377 attested among 417
catalogued for the Indus corpus the statistics reproduce
\citep{mahadevan1977}. No shape was modeled on an Indus sign, and no
shape influenced any statistical result.

\paragraph{The pantheon.}
Shared deity identities combine animal and celestial glosses. The regional
patron assignments emerge from the same finite urn that generates the rest
of the pantheon, with two distinct deities serving the four core regions,
while the two outpost patrons are registered before corpus generation.
Weighted epithet licenses give licensed deities their own collocational
neighborhoods, much as cylinder-seal legends give a deity its recurring
titles \citep{collon1987}. The strongest collocation in
the corpus, Hare-Dawn followed by its licensed epithet Radiant, comes
directly from this mechanism, with a Dunning score of 2,253 against 310.7
for the next pair.

\paragraph{Administrative roles.}
The remaining categories complete the explicit compositional vocabulary.
Guilds can take ranks, commodities can take tallies, issue marks have
bounded service lives, terminals validate an artifact, and ligatures fuse
pairs that were already meaningful, so every statistical regularity in the
corpus can be traced to a relation visible on the plate. The five
terminals alone close 2,126 core texts, and the eight tallies attach to
commodity entries throughout the ledgers. The pairings mirror their
archaeological models: an archaic ledger entry couples a commodity sign
with a count \citep{englund2011}, and a sealing ties an officeholder's
mark to the transaction it validates \citep{zettler1987}.

The vocabulary behind those relations is concrete. The six openers name
auspicious occasions, glossed sunrise, full moon, harvest, high water,
kiln fire, and new year, and the five terminals read as seal emblems,
glossed eye, cup, knot, ladder, and lamp. Thirty of the forty commodities
carry named glosses such as barley, ivory, lapis, and copper, while ten
generic goods numbered 31 through 40 fill out the ledger vocabulary.

The seventeen ligatures are the only identities whose glyphs contain
other glyphs, and composite signs of just this sort are a fixture of the
Indus sign lists themselves \citep{mahadevan1977,parpola1994}. Seven
fuse a commodity with
its tally and five fuse a tally with a terminal, so most fusions live at
the ledger end of a text. One joins two commodities, and the remaining
four pair an opener with a deity, a deity with an epithet, an epithet
with a guild, and a guild with a rank. Every fusion registers a pair that
had already recurred as adjacent signs in the early corpus, and ligature
tokens appear 264 times among the 13,717 core tokens, enough to matter
for segmentation while leaving the component signs in ordinary use.

The office vocabulary follows the same pattern of named and numbered
identities. Half of the 48 guilds carry trade glosses, among them potter,
mason, weaver, smelter, archivist, and water warden, while the other 24
are numbered guilds, and the four ranks grade an office within whichever
guild sign they follow. The twenty issue marks cross the four core
regions with five administrative rounds, glossed sowing allocation, flood
assessment, harvest receipt, dry season redistribution, and year close,
so each mark names a place and an occasion at once. Their bounded service
quotas cap how often any mark can be stamped, 157 uses in all, which
makes them the designed source of the corpus's excess rare types rather
than an accident of sampling.

The two outpost cults, Gulf-Star and Twin-Prows, occupy their own block
of the deity section. They appear only in outpost texts, where Gulf-Star
contributes 77 tokens and Twin-Prows 94, and one or both stand on 168 of
the 350 outpost artifacts. These cults, together with the other
outpost-only deities and epithets, are the unseen events behind the zero
medians of the regional likelihood experiment: a core-trained model has
simply never observed the emblems that carry much of the outpost's
devotional traffic.

The plate also gives the decipherment appendices their concrete
referents. When the Sanskrit decoder reads a brickmaker's tally seal as
a run of Monier-Williams headwords, or the pinned Tamil word list
segments a potter's ledger, the emblems being read are the identities
charted here, and a reader can set any proposed reading beside the
category heading that states what the sign was built to do inside the
invented polity.

\end{document}